\documentclass[letterpaper]{article}

\pdfoutput=1

\usepackage[]{uai2019}
\usepackage[margin=1in]{geometry}
\usepackage{times}

\usepackage{color}
\usepackage{amsmath}
\usepackage{amsbsy}
\usepackage{xspace}
\usepackage{graphicx}
\usepackage{epstopdf}
\usepackage{tikz}
\usetikzlibrary{bayesnet}
\usepackage{amssymb}
\usepackage{subcaption}
\usepackage{multirow}
\usepackage{bm}
\usepackage{boldline}

\usepackage[round]{natbib}

\usepackage[hidelinks]{hyperref}

\title{Deep Gaussian Processes for Multi-fidelity Modeling}

\author{}

\author{
  Kurt Cutajar\\
  EURECOM, France\\
  \href{mailto:cutajar@eurecom.fr}{\text{cutajar@eurecom.fr}}
   \And
    Mark Pullin\\
    Amazon, UK\\
  \href{mailto:marpulli@amazon.com}{\text{marpulli@amazon.com}}
   \And
    Andreas Damianou\\
    Amazon, UK\\
  \href{mailto:damianou@amazon.com}{\text{damianou@amazon.com}}
    \AND
   Neil Lawrence\\
   Amazon, UK\\
  \href{mailto:lawrennd@amazon.com}{\text{lawrennd@amazon.com}}
   \And
   Javier Gonz\'alez\\
    Amazon, UK\\
  \href{mailto:gojav@amazon.com}{\text{gojav@amazon.com}}
}

\begin{document}

%\address{ Institution 1 \And  Institution 2 \And Institution 3 } ]

\definecolor{mygreen}{rgb}{0.2, 0.7, 0.2}
\definecolor{myorange}{rgb}{0.9, 0.5, 0.0}

\newcommand\noteKC[1]{\textcolor{red}{KC - #1}}
\newcommand\noteMP[1]{\textcolor{blue}{MP - #1}}
\newcommand\noteAD[1]{\textcolor{myorange}{AD - #1}}
\newcommand\noteJG[1]{\textcolor{mygreen}{JG - #1}}

\newcommand{\nobs}{n} % Number of observations 
\newcommand{\R}{\mathbb{R}}
\newcommand{\N}{\mathbb{N}}
\newcommand{\Z}{\mathbb{Z}}
\newcommand{\F}{\mathcal{F}}
\newcommand{\I}{\mathcal{I}}
\newcommand{\LL}{\mathcal{L}}
\newcommand{\uu}{\mathbf{u}}
\newcommand{\ee}{\mathbf{e}}

\newcommand{\E}{\mathrm{E}}
\newcommand{\const}{\mathrm{const.}}
\newcommand{\diag}{\mathrm{diag}}
\newcommand{\Tr}{\mathrm{Tr}}

\newcommand{\norm}{\mathcal{N}}

\newcommand{\avect}{\mathbf{a}}
\newcommand{\dvect}{\mathbf{d}}
\newcommand{\fvect}{\mathbf{f}}
\newcommand{\gvect}{\mathbf{g}}
\newcommand{\hvect}{\mathbf{h}}
\newcommand{\mvect}{\mathbf{m}}
\newcommand{\pvect}{\mathbf{p}}
\newcommand{\svect}{\mathbf{s}}
\newcommand{\uvect}{\mathbf{u}}
\newcommand{\vvect}{\mathbf{v}}
\newcommand{\zvect}{\mathbf{z}}
\newcommand{\xvect}{\mathbf{x}}
\newcommand{\yvect}{\mathbf{y}}
\newcommand{\wvect}{\mathbf{w}}
\newcommand{\Wvect}{\mathbf{W}}
\newcommand{\tvect}{\mathbf{t}}
\newcommand{\zerovect}{\mathbf{0}}
\newcommand{\onesvect}{\mathbf{1}}

\newcommand{\betavect}{\boldsymbol{\beta}}
\newcommand{\thetavect}{\boldsymbol{\theta}}
\newcommand{\Thetavect}{\mathbf{\Theta}}
\newcommand{\psivect}{\boldsymbol{\psi}}
\newcommand{\Psivect}{\boldsymbol{\Psi}}
\newcommand{\etavect}{\boldsymbol{\eta}}
\newcommand{\rhovect}{\boldsymbol{\rho}}
\newcommand{\tauvect}{\boldsymbol{\tau}}
\newcommand{\nuvect}{\boldsymbol{\nu}}
\newcommand{\muvect}{\boldsymbol{\mu}}
\newcommand{\omegavect}{\boldsymbol{\omega}}
\newcommand{\Omegavect}{\mathbf{\Omega}}
\newcommand{\sigmavect}{\boldsymbol{\sigma}}
\newcommand{\zetavect}{\boldsymbol{\zeta}}
\newcommand{\varepsilonvect}{\boldsymbol{\epsilon}}
\newcommand{\deltavect}{\boldsymbol{\delta}}

\newcommand{\bigO}{\mathcal{O}}

\newcommand{\name}[1]{{\textsc{#1}}\xspace}

\newcommand{\mcmc}{\name{mcmc}}

% DATASETS
% Small
\newcommand{\cifar}{\name{cifar10}}
\newcommand{\mnisteight}{\textsc{mnist}8\textsc{m}\xspace}
\newcommand{\protein}{\name{protein}}
\newcommand{\powerplant}{\name{powerplant}}
\newcommand{\spam}{\name{spam}}
\newcommand{\eeg}{\name{eeg}}
\newcommand{\usps}{\name{usps}}
% Medium
%% \newcommand{\sarcos}{\name{sarcos}} % sarcos all
\newcommand{\mnist}{\name{mnist}} % mnist all
%% \newcommand{\mnistbin}{\name{mnist-b}}
%% \newcommand{\sarcostwo}{\name{sarcos-2}}
% Large
\newcommand{\airline}{\name{airline}}

% Models
\newcommand{\autogp}{\textsc{a}uto\textsc{gp}\xspace}
\newcommand{\mfdgp}{\textsc{mf-dgp}\xspace}
\newcommand{\nargp}{\textsc{nargp}\xspace}
\newcommand{\deepmf}{\textsc{deep-mf}\xspace}
\newcommand{\ar}{\textsc{ar{\scriptsize 1}}\xspace}
\newcommand{\dmf}{\textsc{deep-mf}\xspace}
\newcommand{\dsdgp}{\textsc{ds-dgp}\xspace}
\newcommand{\gpflow}{\textsc{GP}flow\xspace}
\newcommand{\tensorflow}{\textsc{T}ensor\textsc{F}low\xspace}

\newcommand{\gp}{\name{gp}}
\newcommand{\svi}{\name{svi}}
\newcommand{\gps}{\textsc{gp}s\xspace}
\newcommand{\dgp}{\name{dgp}}
\newcommand{\dgps}{\textsc{dgp}s\xspace}
\newcommand{\ard}{\name{ard}}

\newcommand{\rbf}{\name{rbf}}
\newcommand{\dpp}{\name{dpp}}

% Metrics
\newcommand{\mnll}{\name{mnll}}
\newcommand{\rmse}{\name{rmse}}
\newcommand{\nelbo}{\name{nelbo}}

\maketitle

\begin{abstract}
	%!TEX root = main.tex
%Multi-fidelity methods are prominently used when cheaply-obtained, but possibly biased and noisy observations must be effectively combined with limited or expensive true data in order to construct reliable models.
%This arises in both fundamental machine learning procedures such as Bayesian optimization, as well as more practical science and engineering applications.
%The appeal of applying deep Gaussian processes (\dgps) to this setting rests on capturing complex nonlinear correlations across fidelities.
%However, the architectures explored thus far are burdened by structural assumptions and constraints which deter the flexibility of such models.
%In this paper we propose a novel multi-fidelity model which treats \dgp layers as fidelity levels and uses a variational inference scheme to propagate uncertainty across them.
%We show that this approach makes substantial improvements in quantifying and propagating uncertainty in multi-fidelity set-ups, which in turn improves their effectiveness in decision-making pipelines.

% % INCORPORATING ANDREAS' FEEDBACK

Multi-fidelity methods are prominently used when cheaply-obtained, but possibly biased and noisy, observations must be effectively combined with limited or expensive true data in order to construct reliable models.
This arises in both fundamental machine learning procedures such as Bayesian optimization, as well as more practical science and engineering applications.
In this paper we develop a novel multi-fidelity model which treats layers of a deep Gaussian process as fidelity levels, and uses a variational inference scheme to propagate uncertainty across them.
This allows for capturing nonlinear correlations between fidelities with lower risk of overfitting than existing methods exploiting compositional structure, which are conversely burdened by structural assumptions and constraints.
We show that the proposed approach makes substantial improvements in quantifying and propagating uncertainty in multi-fidelity set-ups, which in turn improves their effectiveness in decision making pipelines.
\end{abstract}

\section{Introduction}

\begin{figure*}[t!]
	\centering
	\includegraphics[width=.5\textwidth]{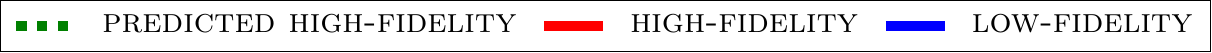}
	\setlength{\tabcolsep}{1pt}
	\begin{tabular}{ccc}
		\ar &\nargp& \mfdgp \vspace{-.1cm}\\
		\includegraphics[width=155pt]{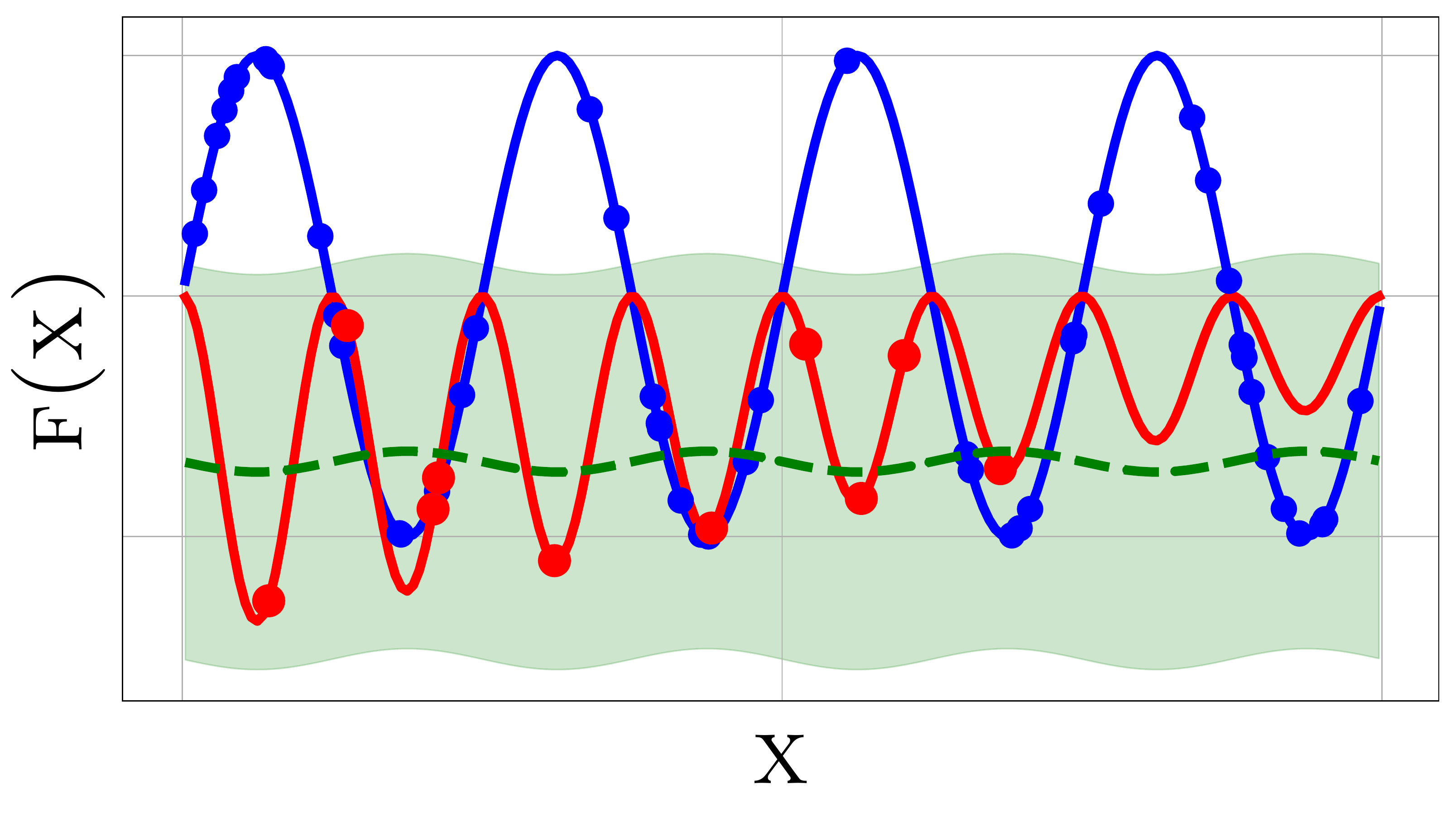} &
		\includegraphics[width=155pt]{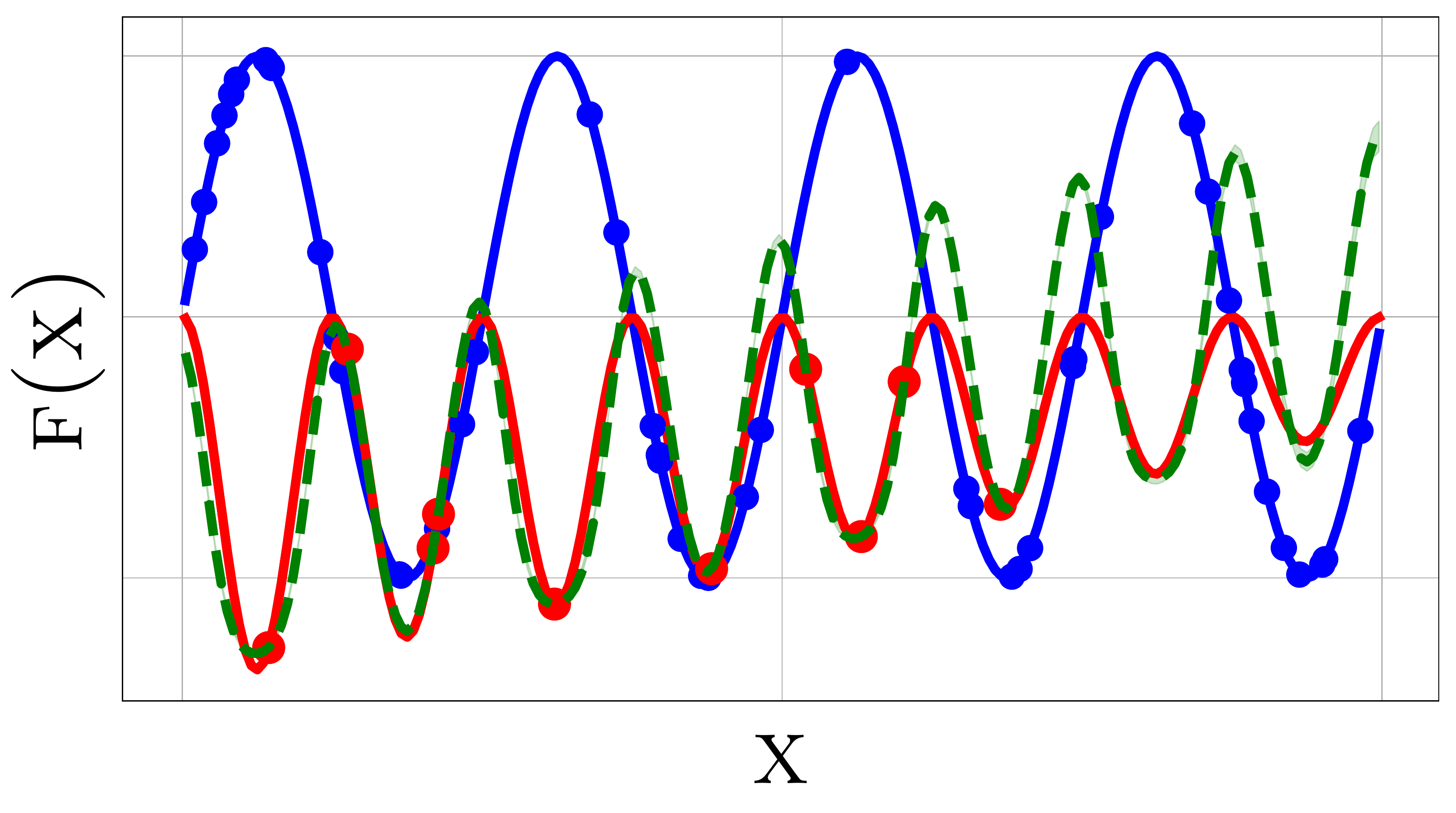}&
		\includegraphics[width=155pt]{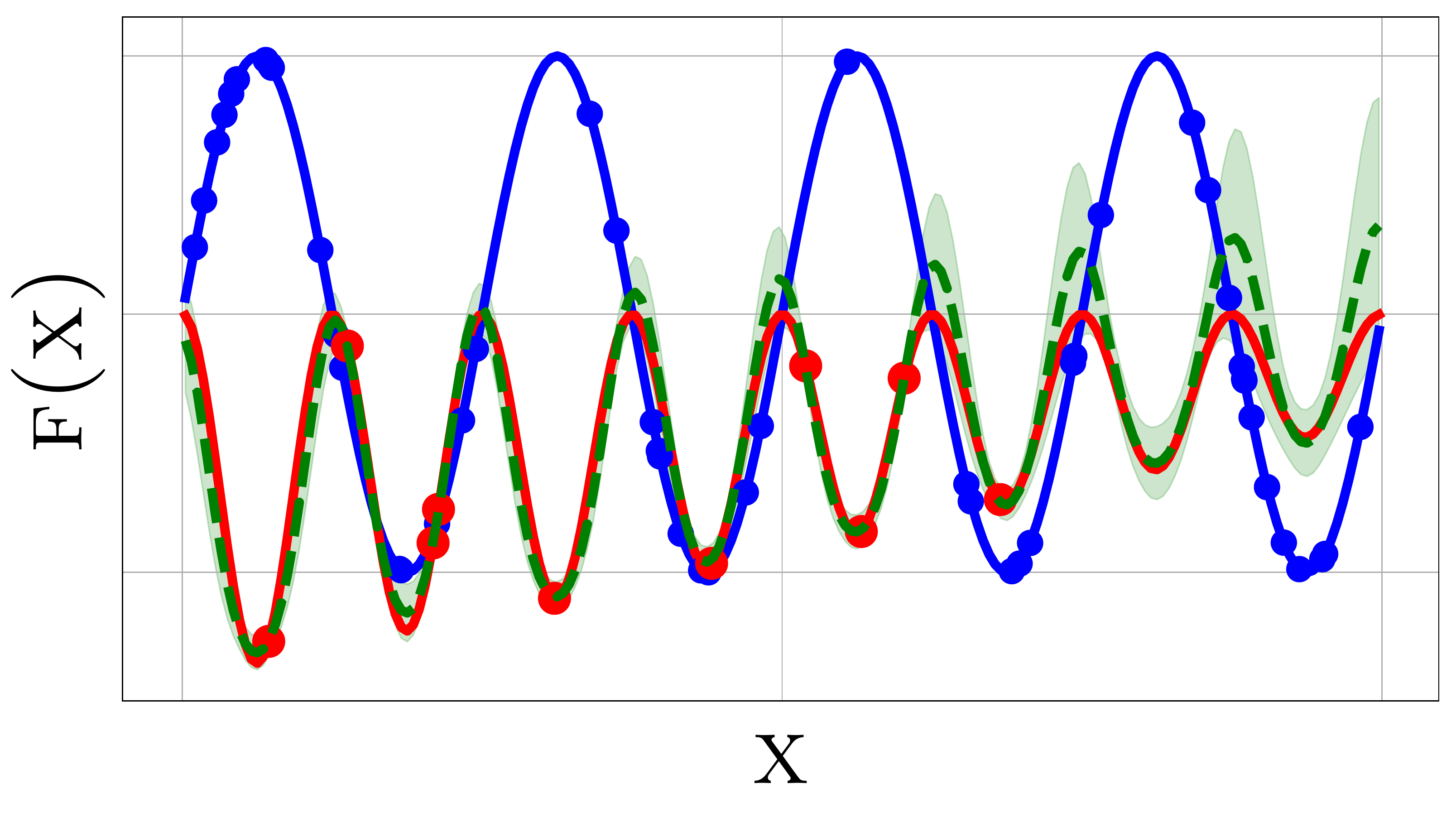} 
	\end{tabular}
	\vspace*{-5.5mm}
	\caption{Limitations addressed and resolved jointly by our proposed \mfdgp architecture.
		Blue and red markers denote low and high-fidelity observations respectively.
		Shaded regions indicate the 95\% confidence interval.}
	\label{fig:intro_model_comp}
\end{figure*}

A common issue encountered in active learning procedures such as Bayesian optimization \citep{Shahriari2016} and experimental design \citep{Morris2004} is the difficulty or cost to acquire sufficient data.
Constructing a reliable model of the underlying system when only few observations are available is challenging, making it common practice to develop simulators from which data can more easily be obtained.
Practical examples include computational fluid dynamics for vehicular engineering~\citep{Koziel2013}, weather simulators for climate modeling~\citep{Majda2010}, and emulators for reinforcement learning \citep{Cutler2014}.

Multi-fidelity models~\citep{Kennedy2000,Peherstorfer2018} are designed to fuse limited true observations (\textit{high-fidelity}) with cheaply-obtained lower granularity approximations (\textit{low-fidelity}).
%In such models, data acquired from the true source is referred to as \textit{high-fidelity}, whereas lower granularity representations are denoted as being \textit{low-fidelity}.
However, na\"ively combining data from multiple information sources could result in a model giving predictions which do not accurately reflect the true underlying system.
In absence of well-defined information regarding the reliability of each fidelity and the relationships between fidelities, Bayesian inference captures the principle of Occam's razor through explicitly encoding our uncertainty about these factors~\citep{Mackay2003}.
%the principle of Occam's razor by encoding uncertainty within the model's formulation~\citep{Mackay2003}.
This implicit regularization is pertinent to settings with limited data where overfitting is otherwise likely to occur.

In the spirit of Bayesian modeling, Gaussian processes~\citep[\gps;][]{Rasmussen2006} are well suited to multi-fidelity problems due their ability to encode prior beliefs about how fidelities are related, yielding predictions accompanied by uncertainty estimates.
\gps formed the basis of seminal autoregressive models (henceforth \ar) investigated by~\cite{Kennedy2000} and \cite{LeGratiet2014}, and were shown to be effective given a linear mapping between fidelities, i.e.\ the high-fidelity function $f_t$ can be modeled as:
\begin{equation}
f_t(x) = \rho_t f_{t-1}\left(x\right) + \delta_t\left(x\right) \text{,}
\end{equation}
where $\rho_t$ is a constant scaling the contribution of samples $f_{t-1}$ drawn from the \gp modeling the data at the preceding fidelity, and $\delta_{t}(x)$ models the bias between fidelities.
However, such models are insufficient when the relationship between fidelities is nonlinear, i.e.\ there is now a space-dependent nonlinear transformation $\rho_t$ that relates one fidelity to the next as:
\begin{equation}
f_t(x) = \rho_t\left(f_{t-1}\left(x\right), x\right) + \delta_t\left(x\right).
\end{equation}
The additive structure and independence assumption between the \gps for modeling $\rho_t\left(f_{t-1}\left(x\right), x\right)$ and $\delta_t\left(x\right)$ permits us to combine these as a single \gp that takes as inputs both $x$ and $f^*_{t-1}(x)$, where the latter denotes a sample from the posterior of the \gp modeling the preceding fidelity evaluated at $x$.
This results in a composition of \gps that can be compactly expressed as $f_t(x) = g_t\left(f_{t-1}^*\left(x\right), x\right)$.
As highlighted by \cite{Perdikaris2017} and exemplified in Figure~\ref{fig:intro_model_comp}, the \ar model cannot capture nonlinear correlations between fidelities.

\subsection*{Problem Statement}

Deep Gaussian processes~\citep[\dgps;][]{Damianou2013} are a natural candidate for handling nonlinearities between fidelities by way of function composition, allowing for uncertainty propagation in a nested structure of \gps where each \gp models the transition from one fidelity to the next.
However, \dgps are cumbersome to develop and approximations are necessary for enabling tractable inference.
In spite of being motivated by the structure of \dgps, the nonlinear multi-fidelity model (\nargp) proposed by~\cite{Perdikaris2017} amounts to a disjointed architecture whereby each \gp is fitted in an isolated hierarchical manner, thus preventing \gps at lower fidelities from being updated once they have already been fit.
This deconstruction into independent models which are optimized sequentially violates our aforementioned preference of using Occam's razor as a means of controlling the model's complexity, making it more susceptible to overfitting.
 
Consider the example given in Figure~\ref{fig:intro_model_comp}.
In the tail-end of the function, there are no high-fidelity observations available and only low-fidelity points to fall back on. 
In this case, we would expect the model to return higher uncertainty to reflect the lack of data available, but instead, \nargp predicts an incorrect result with high confidence.
Closer inspection of the optimal hyperparameters obtained after training the model confirms our intuition regarding overfitting, since kernel parameters settle at values which are orders of magnitude larger than the range in which they are expected to lie.
This is particularly problematic when the model is intended for use in a computational pipeline or active learning procedure, where uncertainty calibration is imperative.

In this work, we propose the first complete interpretation of multi-fidelity modeling using \dgps, which we refer to as \mfdgp.
In particular, we construct a multi-fidelity \dgp model which can be trained end-to-end, overcoming the constraints that hinder existing attempts at using \dgp structure for this purpose.
Having a \dgp model that communicates uncertainty estimates between all fidelities at training time also allows us to properly assess the suitability of \dgps over standard \gps in the multi-fidelity setting.
Returning to the example given in Figure~\ref{fig:intro_model_comp}, we see that our model fits the true function properly while also returning sensibly conservative uncertainty estimates.
Moreover, our model also inherits the compositional structure of \nargp, thus alleviating a crucial limitation of the \ar model.
The model's formulation leverages the sparse \dgp approximation proposed by~\cite{Salimbeni2017} for tractability.

Our principal contributions are listed below:
\begin{itemize}
	\item We identify potential issues with existing approaches for compositional multi-fidelity modeling, emphasising their tendency to overfit;
	\item We develop a novel multi-fidelity model which enables end-to-end training with well-calibrated uncertainty quantification.
	This includes a detailed analysis of the nuances involved in its construction;
	\item We provide a thorough experimental evaluation of our model by way of comparisons with other techniques, application to a large-scale real-world problem, and showcase the use of \mfdgp for experimental design using a determinantal point process;
	\item The model implementation has been integrated in Emukit\footnote{\href{https://github.com/amzn/emukit}{\text{https://github.com/amzn/emukit}}}, an open-source package for carrying out emulation and decision making in a design loop.
\end{itemize}

The paper is organized as follows.
In the next section, we review the literature on multi-fidelity modeling with \gps and clarify how our contributions fit within this landscape.
Subsequently, in Section~\ref{sec:dgps} we introduce \dgps and illustrate how these can be interpreted in the multi-fidelity setting.
A detailed description and discussion of our model, \mfdgp, follows in Section~\ref{sec:mf_dgp}, and its performance is evaluated in Section~\ref{sec:experiments}, where we also compare our results against a selection of alternatives.
An outlook on extensions and future work concludes the paper.
\section{Related Work}\label{sec:related_work}

Multi-fidelity models came to prominence in the foundational work by \citet{Kennedy2000}, where a \gp having a kernel suited for multi-fidelity observations was used to model linear correlations between data at $T$ ordered fidelity levels.
However, the flexibility of this approach was burdened by the cubic computational complexity associated with \gp inference.
This led \cite{LeGratiet2014} to propose a recursive multi-fidelity model whereby the observations for each fidelity are modeled using independent \gps.
Aside from reducing the computational complexity from $\bigO((\sum_{t=1}^{T}N_t)^3)$ to $\bigO(\sum_{t=1}^{T}N_t^3)$, where $N_t$ denotes the number of observations with fidelity level $t$, the posterior obtained from this model was also shown to be identical to that of the original model under the assumptions of noiseless observations and nested inputs, i.e.\ points observed with fidelity level $t$ are also observed at all lower fidelities.

The similarity between nested \gp models for multi-fidelity and traditional deep \gps was first noted by~\cite{Perdikaris2017} in their formulation of the \nargp model, where the parallels to \dgp inference are derived from propagating uncertain outputs from one \gp to the next.
Nonetheless, the design and implementation of our \mfdgp model is markedly different, and this has notable implications on both the model architecture as well as its predictive performance.
Whereas \nargp amounts to a set of disjointed \gps trained sequentially in isolation, here we present a single \dgp for jointly modeling data from all fidelities; \nargp disregards the nuances of such models in its formulation.

%in spite of the parallels to \dgp inference derived from propagating uncertain outputs from one \gp to the next, the \nargp model proposed there requires each \gp to be trained independently, while also assuming nested and noiseless observations for reliable inference.
%The model developed in our work lifts these constraints by instead constructing a single \dgp model which can be trained end-to-end.
%, thus ensuring that uncertainty is propagated in a principled manner without overfitting to observations at lower layers.
Conversely, the `deep multi-fidelity \gp' model (\deepmf) presented by~\cite{Raissi2016} extends the original multi-fidelity model by learning a deterministic transformation applied to the inputs (using a deep neural network).
However, the resulting model bears more resemblance to a manifold \gp~\citep{Calandra2016}, which amounts to standard \gp inference on warped inputs and does not involve actual process composition.
The autoregressive nature of \dgps is also briefly mentioned in~\cite{Requeima2018}.

Several other extensions to traditional multi-fidelity approaches have been developed, singularly addressing issues such as scalability~\citep{Zaytsev2017}, mismatched training and target distributions~\citep{Liu2018}, incorporating gradient information~\citep{Ulaganathan2016}, and non-hierarchical ordering of fidelities~\citep{Lam2015, Poloczek2017}.
Tangentially, multi-fidelity methods tailored to Bayesian optimization and bandit algorithms have also recently been investigated by~\cite{Sen2018} and~\cite{Kandasamy2016} among others.
%A variety of application-specific model variations have also appeared in various engineering applications.
\section{Deep Gaussian Processes}\label{sec:dgps}

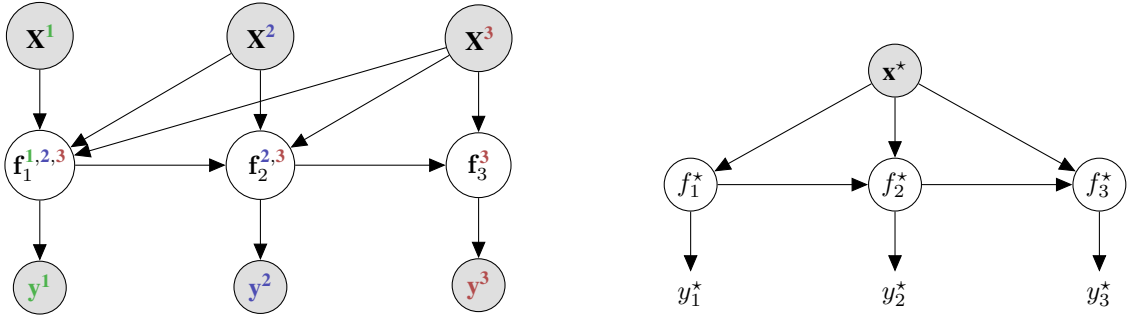
\begin{figure*}[t]
	\centering
	\begin{tabular}{cc}
		\begin{tikzpicture}

		\node[latent]  (FL) 	{{$ \hphantom{a}\fvect_3^{\textcolor{red!40!gray}{\textbf{3}}}$\hphantom{b}}};
		\node[latent, left=2 of FL]  (F2) 		{{$ \hphantom{a}\fvect_2^{\textcolor{blue!40!gray}{\textbf{2}},\textcolor{red!40!gray}{\textbf{3}}}$}};
		\node[latent, left=2 of F2]  (F1) 		{{$\fvect^{\textcolor{green!40!gray}{\textbf{1}},\textcolor{blue!40!gray}{\textbf{2}},\textcolor{red!40!gray}{\textbf{3}}}_1$}};
		
		\node[obs, above=.8 of F1]		(X1) 	{$\hphantom{a}\textbf{X}^{\textcolor{green!40!gray}{\textbf{1}}}\hphantom{b}$};
		\node[obs, above=.8 of F2]		(X2) 	{$\hphantom{a}\textbf{X}^{\textcolor{blue!40!gray}{\textbf{2}}}\hphantom{b}$};
		\node[obs, above=.8 of FL]		(XL) 	{$\hphantom{a}\textbf{X}^{\textcolor{red!40!gray}{\textbf{3}}}\hphantom{b}$};
		
		\node[obs, below=.8 of F1]		(y1) 	{$\textcolor{green!40!gray}{\textbf{y}^\textbf{1}}$};
		\node[obs, below=.8 of F2]		(y2) 	{$\textcolor{blue!40!gray}{\textbf{y}^\textbf{2}}$};
		\node[obs, below=.8 of FL] 		(yL)	{$\textcolor{red!40!gray}{\textbf{y}^\textbf{3}}$};
		
%		\edge[->] {X2} {F2};
%		\edge[->] {XL} {FL};
		
		\edge[->] {F1} {y1} ;
		\edge[->] {F2} {y2} ;
		\edge[->] {FL} {yL} ;
		
		\edge[->] {F1} {F2} ;
		\edge[->] {F2} {FL} ;
		
		\edge[->] {X1}{F1}
		\edge[->] {X2}{F2}
		\edge[->] {X2}{F1}
		\edge[->] {XL}{FL}
		\edge[->] {XL}{F2}
		\edge[->] {XL}{F1}
		
\end{tikzpicture}
\vspace{.3cm}\hspace{1.5cm} &
		\begin{tikzpicture}

		\node[latent]  (FL) 	{{$f_3^{\star}$}};
		\node[latent, left=2 of FL]  (F2) 	{{$f_2^{\star}$}};
		\node[latent, left=2 of F2]  (F1) 	{{$f_1^{\star}$}};
		
%		\node[obs, above=.8 of F1]		(X1) {{$\textbf{x}^{\star}$}};
		\node[obs, above=.8 of F2]		(X2) {{$\textbf{x}^{\star}$}};
%		\node[obs, above=.8 of FL]		(XL) {{$\textbf{x}^{\star}$}};
		
		\node[below=.8 of F1]		(y1) 	{${y}_1^\star$};
		\node[below=.8 of F2]		(y2) 	{${y}_2^\star$};
		\node[below=.8 of FL] 		(yL)	{${y}_3^\star$};
		
%		\edge[->] {X2} {F2};
%		\edge[->] {XL} {FL};
		
		\edge[->] {F1} {y1} ;
		\edge[->] {F2} {y2} ;
		\edge[->] {FL} {yL} ;
		
		\edge[->] {F1} {F2} ;
		\edge[->] {F2} {FL} ;
		
		\edge[->] {X2}{F1}
		\edge[->] {X2}{F2}
		\edge[->] {X2}{FL}
		
\end{tikzpicture}
\vspace{.3cm}
		% % ORIGINAL DIAGRAMS BELOW
%		\input{figures/dgp_mf_arch_3l}\hspace{1.5cm} &
%		\input{figures/dgp_mf_pred_3l}
	\end{tabular}
	\vspace*{-5mm}
	\caption{\textit{Left:} \mfdgp architecture with three fidelity levels.
	Observed data and latent variables are color-coded in order to indicate the associated fidelity level.
	The latent variables at each layer denote samples drawn from a \gp.
	For example, the evaluation of \mfdgp at layer `1' for the inputs observed with fidelity `3' is denoted as $\fvect_1^3$.
	 \textit{Right: } Predictions using the same \mfdgp model, whereby the original input $\xvect_\star$ is input at every fidelity level along with the evaluation up to the previous level.
	 The output $y_t^\star$ denotes the model's prediction for fidelity $t$.}
	\label{fig:mfdgp_3l}
\end{figure*}

Consider a supervised learning problem in which we are interested in learning the mapping between a set of $N$ input vectors $\textbf{X} = [\xvect_1, \ldots, \xvect_N]^{\top}$, where $\xvect_i \in \mathbb{R}^{D_\text{in}}$, and corresponding univariate labels $\yvect = [y_1, \ldots, y_N]^{\top}$, with $y_i \in \mathbb{R}$.
Gaussian processes \citep[\gp{s};][]{Rasmussen2006} rely on Bayesian inference for learning a mapping such that the distribution over any finite subset of input points is a multivariate Gaussian.
More formally, observations are assumed to be noisy realisations of function values $\fvect = [f_1, \ldots, f_N]^{\top}$ drawn from a \gp with some likelihood $p(\yvect | \fvect)$.
%Consider a supervised learning problem where a set of $n$ input vectors $X = [\xvect_1, \ldots, \xvect_\nobs]^{\top}$ is associated with a set of univariate labels $\yvect = [y_1, \ldots, y_\nobs]^{\top}$, where $\xvect_i \in \mathbb{R}^{D_\text{in}}$ and $y \in \mathbb{R}$.
%We are interested in characterizing a mapping between the inputs $X$ and the corresponding labels $\yvect$, for which we can apply Gaussian processes \citep[\gp{s};][]{Rasmussen2006}.
%\gp models assume that observations are distributed according to a distribution $p(y_i | f_i)$ and that $\fvect = [f_1, \ldots, f_\nobs]^{\top}$ is distributed as a \gp. 
The key characteristics of the functions that can be drawn from the \gp are determined by a set of covariance parameters defining the \gp prior.
A popular choice of covariance is the exponentiated quadratic (or \rbf) function:
\begin{equation}\label{eqn:rbf}
k(\xvect_i, \xvect_j | \thetavect) = 
\sigma^2 \exp\left[-\frac{1}{2} (\xvect_i - \xvect_j)^{\top} \Lambda^{-1} (\xvect_i - \xvect_j) \right] \text{,}
\end{equation}

where the parameter set $\thetavect$ comprises the marginal variance of the \gp, $\sigma^2$, and $\Lambda = \diag(l_1^2, \ldots, l_{D_\text{in}}^2)$, with each $l_d$ interpreted as a lengthscale parameter.
The posterior distribution of a \gp denotes a Gaussian distribution over candidate functions characterized by a posterior mean and covariance.

Inspired by the widespread success of deep learning in neural network architectures, deep Gaussian processes \citep[\dgp{s};][]{Damianou2013} are constructed by nesting \gp models such that the output of one \gp is propagated as input to the next.
Their application to the multi-fidelity setting is particularly appealing because if we assume that each layer corresponds to a fidelity level, then the latent functions at the intermediate layers are given a meaningful interpretation which is not always available in standard \dgp models.

However, in spite of their theoretic appeal, inference using \dgp models is notoriously difficult since the integrals involved in computing the marginal likelihood and making predictions are generally intractable \citep{Damianou2015}.
The first attempt at using \dgp structure in a multi-fidelity setting~\citep{Perdikaris2017} relied on structural assumptions on the data to circumvent these difficulties.
However, the model's capacity and flexibility are heavily impaired by such simplifications.

Recent advances in the \dgp literature~\citep{Cutajar2017,Salimbeni2017} have leveraged traditional \gp approximations to construct scalable \dgp models which are easier to specify and train.
While both of the aforementioned \dgp approximations can be adapted for multi-fidelity data, we peruse the model presented by~\cite{Salimbeni2017} to avoid the constraints imposed on selecting kernel functions in~\cite{Cutajar2017}.
\section{Multi-fidelity DGP (\mfdgp)}\label{sec:mf_dgp}

Extending the concepts introduced in the previous section, we now describe the architecture of our proposed \mfdgp model along with the nuances of its design.
In the spirit of continuity, we intentionally mirror the notation of~\cite{Salimbeni2017} to preserve focus on the components enabling multi-fidelity modeling.

\subsection{Model Specification}

Let us assume a dataset $\mathcal{D}$ having observations at $T$ fidelities, where $\textbf{X}^t$ and $\textbf{y}^t$ denote the inputs and corresponding outputs observed with fidelity level $t$:
\begin{equation*}
\mathcal{D} = \left\{\left(\textbf{X}^1, \textbf{y}^1\right), \dots, \left(\textbf{X}^t, \textbf{y}^t\right), \dots, \left(\textbf{X}^T, \textbf{y}^T\right) \right\}.
\end{equation*}

Intuitively, and for enhanced interpretability, we assume that each layer of our \mfdgp model corresponds to the process modeling the observations available at fidelity level $t$, and that the bias or deviation from the true function decreases from one level to the next.
We use the notation $\fvect_l^t$ to denote the evaluation at layer $l$ for inputs observed with fidelity $t$; for example, the evaluation of the process at layer `1' for the inputs observed with fidelity `3' is denoted as $\fvect^3_1$.
A conceptual illustration of the proposed \mfdgp architecture is given in Figure~\ref{fig:mfdgp_3l} (\textit{left}) for a dataset with three fidelities.
Note that the \gp at each layer is conditioned on the data belonging to that level, as well as the evaluation of that same input data at the preceding fidelity.
This gives an alternate perspective to the notion of feeding forward the original inputs at each layer, as originally suggested in \cite{Duvenaud2014} for avoiding pathologies in deep architectures.

\subsection*{Layer-wise Sparse Approximation}

At each layer we rely on the sparse variational approximation of a \gp for inference, whereby a set of inducing points $\uvect$ is introduced such that the augmented joint posterior $p\left(\fvect, \uvect\right)$ yields a true bound on the marginal likelihood of the exact GP.
This is achieved by introducing:

\begin{equation}\label{eqn:svi_full}
q\left(\fvect^t_l | \uvect_l\right) =
p\left( \fvect_l^t | \uvect_l; \{\fvect^t_{l-1}, \textbf{X}^t\}, \textbf{Z}_{l-1} \right)q\left(\uvect_l\right),
\end{equation}

where $\mathbf{Z}_{l-1}$ denotes the inducing inputs for layer $l$, $\uvect_l$ their corresponding function evaluation, and $q\left(\uvect_l\right) = \mathcal{N}\left(\uvect_l| \pmb{\mu}_l, \pmb{\Sigma}_l\right)$ is the variational approximation of the inducing points.
The mean and variance defining this variational approximation, i.e.\ $\pmb{\mu}_l$ and $\pmb{\Sigma}_l$, are optimized during training.
Furthermore, if $\uvect_l$ is marginalized out from Equation~\ref{eqn:svi_full}, the resulting variational posterior is once again Gaussian and fully defined by its mean, $\widetilde{\mathbf{m}}_l$, and variance, $\widetilde{\mathbf{S}}_l$:

\begin{equation}\label{eqn:svi_intU}
q\left(\fvect^t_l | \pmb{\mu}_l, \pmb{\Sigma}_l; \{ \fvect_{l-1}^t, \textbf{X}^t\}, \textbf{Z}_{l-1} \right) = \mathcal{N}\left(\fvect^t_l \;|\; \widetilde{\mathbf{m}}_l^t, \widetilde{\mathbf{S}}_l^t\right),
\end{equation}

which can be derived analytically.
%Again, we refer the reader to the work by \cite{Salimbeni2017} for more detail on the \dgp approximation upon which our model is based.

The marginalization property which is key to simplifying inference is also preserved in the multi-fidelity setting.
In particular, this entails that within each layer the marginals depend exclusively on the corresponding inputs, yielding the following posterior for the $i^{\text{th}}$ input observed with highest fidelity:
\begin{align}\label{eqn:svi_marginal}
 q\left( {f}^{i,T}_{L} \right) =  \int & \prod_{l=1}^L \left[ q\left({f}^{i,T}_{l} | \pmb{\mu}_l, \pmb{\Sigma}_l ; \left\lbrace{f}^{i,T}_{l-1}, \textbf{x}^{i,T} \right\rbrace, \textbf{Z}_{l-1}  \right) \right] \nonumber \\ &  \text{d}{f}^{i,T}_{1} \dots \text{d}{f}^{i,T}_{L - 1}.
\end{align}

%As with other \dgp models~\citep{Dai2016, Cutajar2017} trained using stochastic variational inference (see Section~\ref{sec:svi}), the reparameterization trick~\citep{Kingma2014} is then used to recursively draw samples from the variational posterior:
%
%\begin{align}
%{\hat{f}}^{i,t}_{l} = &\;\tilde{\mathbf{m}}_l \left(\left\lbrace{\hat{f}}^{i,t}_{l-1}, \textbf{x}^{i,t} \right\rbrace \right) + \nonumber\\
%&\;\mathbf{\varepsilon}^{i,t}_l \odot \sqrt{\tilde{\mathbf{S}}_l \left(\left\lbrace{\hat{f}}^{i,t}_{l-1},\textbf{x}^{i,t} \right\rbrace, \left\lbrace{\hat{f}}^{i,t}_{l-1},\textbf{x}^{i,t}\right\rbrace \right) }\;\;,
%\end{align}
%
%where $\mathbf{\varepsilon}^{i,t}_l\sim \mathcal{N}\left( \textbf{0}, \textbf{I}_{D_{\text{out}}} \right)$.

Note that at all layers, $\uvect_l$ will have dimensionality $M_l \times D_{\text{out}}$, where $M_l$ is the number of inducing points at layer $l$ and $D_\text{out}$ is the output dimensionality of the observations.
On the other hand, $\mathbf{Z}_{l-1}$ will have dimensionality $M_l \times D_{\text{in}}$ at the first layer, but $M_l \times \left(D_{\text{in}} + D_{\text{out}}\right)$ at all subsequent ones.
This happens because the intermediate layers' inputs contain both the location of the data point in the original input space as well as its evaluation up to the previous layer/fidelity.
The likelihood noise at lower fidelity levels is encoded as additive noise in the kernel function of the \gp at that layer.

\subsection*{Evidence Lower Bound}

We can formulate the variational lower bound on the marginal likelihood as follows:
\begin{align}\label{eqn:marg_lik}
	\mathcal{L}_\mfdgp &= \sum_{t=1}^T\sum_{i=1}^{N_t}\mathbb{E}_{q\left({f}^{i,t}_t\right)}\left[ \log p\left({y}^{i,t} | {f}^{i,t}_t\right) \right] \nonumber \\ 
	& + \sum_{l=1}^L D_\text{KL}\left[ q\left(\uvect_l\right) ||\;p\left(\uvect_l; \mathbf{Z}_{l-1}\right)\right],
\end{align}
where we assume that the likelihood is factorized across fidelities and observations (allowing us to express the log likelihood as a double summation), and $D_\text{KL}$ denotes the Kullback-Leibler divergence.
This lower bound is the multi-fidelity objective function for our model, and a full derivation can be found in the supplementary material.

\subsection{Multi-fidelity Predictions}

Model predictions with different fidelities are also obtained recursively by propagating the input through the model up to the chosen fidelity.
At all intermediate layers, the output from the preceding layer `$t$-1' (also corresponding to the prediction with fidelity `$t$-1') is augmented with the original input, as will be made evident by the choice of kernel explained in the next section.
The output of a test point $\textbf{x}^\star$ can then be predicted with fidelity level $t$ as follows:

\begin{equation}\label{eqn:prediction}
q \left({f}^{\star}_t \right) \approx \frac{1}{S}\sum^{S}_{s=1} q\left( {f}^{s,\star}_t | \pmb{\mu}_t, \pmb{\Sigma}_t ; \{{f}^{s,\star}_{t-1}, \textbf{x}^\star \}, \textbf{Z}_{t-1}\right),
\end{equation}

where $S$ denotes the number of Monte Carlo samples and $t$ replaces $l$ as the layer indicator (assuming one layer per fidelity).
This procedure is illustrated in Figure~\ref{fig:mfdgp_3l} (\textit{right}).

\subsection{Multi-fidelity Covariance}\label{sec:kernel_design}

The multi-fidelity kernel function for every \gp at an intermediate layer is inspired by that proposed in \cite{Perdikaris2017}, since it captures both the potentially nonlinear mapping between outputs as well as the correlation in the original input space:

\begin{align}\label{eqn:composite_kernel}
k_l = &\;k_l^\rho\left(\textbf{x}^i, \textbf{x}^j; \pmb{\theta}_l^\rho\right) k_l^{f-1}\left({f}^*_{l-1}(\textbf{x}^i), {f}^*_{l-1}(\textbf{x}^j); \pmb{\theta}_l^{f-1}\right) \nonumber\\	
&+ k_l^\delta\left(\textbf{x}^i, \textbf{x}^j; \pmb{\theta}_l^\delta\right) \text{,}
\end{align}

where $ k_l^{f-1}$ denotes the covariance between outputs obtained from the preceding fidelity level, $ k_l^\rho$ is a space-dependent scaling factor, and $k_l^\delta$ captures the bias at that fidelity level.
At the first layer this reduces to:

\begin{equation}\label{eqn:l0_kernel}
k_1 = k_1^\delta\left(\textbf{x}^i, \textbf{x}^j; \pmb{\theta}_1^\delta\right).
\end{equation}

\cite{Perdikaris2017} assumed that each individual component of the composite kernel function is an \rbf kernel as defined in Equation~\ref{eqn:rbf}; however, this may not be appropriate when the mapping between fidelities is linear.
To this end, we propose to enhance the covariance function given in Equation~\ref{eqn:composite_kernel} with a linear kernel such that the composite intermediate layer covariance becomes:
\begin{align}\label{eqn:composite_kernel_linear}
k_l = &\;k_l^\rho\left(\textbf{x}^i, \textbf{x}^j; \pmb{\theta}_l^\rho\right) \left[
\vphantom{ k_l^{f-1}\left({f}_{l-1}(\textbf{x}^i), {f}_{l-1}(\textbf{x}^j); \pmb{\theta}_l^{f-1}\right)}
\sigma^2_l{f}^*_{l-1}(\textbf{x}^i)^\top {f}^*_{l-1}(\textbf{x}^j) \right. \nonumber\\	
&\left. + \;k_l^{f-1}\left({f}^*_{l-1}(\textbf{x}^i), {f}^*_{l-1}(\textbf{x}^j); \pmb{\theta}_l^{f-1}\right)\right]\nonumber\\
&+\;k_l^\delta\left(\textbf{x}^i, \textbf{x}^j; \pmb{\theta}_l^\delta\right).
\end{align}
A similar discussion on designing more tailored kernels for autoregressive problems was recently also put forward by~\cite{Liu2018} and~\cite{Requeima2018}.

%\noteKC{Could be very cool to show a comparison of the kernel matrices in this section.
%One for high-fidelity only, and then layer 1 + layer 2. DECIDED TO POSTPONE THIS ONE.}

\subsection{Treatment of Inducing Inputs}\label{sec:indudcing_inputs}

One of the less straightforward aspects of this model concerns the selection and optimization of inducing inputs at \textit{}layers 2 to \textit{L}.
Recall that the first layer only takes input points lying in the standard input space of the function; in this case, the role of inducing inputs is straightforward as in other sparse \gp approximations.
However, the transition to higher layers is not as clear.

At these layers, the input to the intermediate \gp is the combination of points in the original input space as well as the corresponding function evaluation returned from the previous layer.
However, freely optimizing inducing points at these layers is no longer appropriate since the output from the previous layer is intrinsically linked to the input point with which it is associated.
We currently circumvent this issue by selecting the inducing points from the available observations at the previous fidelity layer and fix them during optimization.
Devising more principled approaches for constraining the optimization of inducing points is a challenging direction for future work.

%Several papers have noted that sparse approximations can be used to extend their proposed multi-fidelity models to larger datasets.
%%However, this aspect has never yet been properly explored in practice, along with all the unexpected nuances such approximations may present in the context of multi-fidelity problems.
%The low-rank Nystr\"om approximation was briefly considered by~\cite{Zaytsev2017}, but that investigation was limited to the variable fidelity setting (two levels), and does not generalize beyond the standard assumption of linear mappings between fidelities.

\subsection{Stochastic Variational Inference}\label{sec:svi}

The use of stochastic variational inference (\svi) techniques~\citep{Hoffman2013, Hensman2014} for optimizing kernel parameters and inducing inputs requires careful design for ensuring the model consistently converges to an optimal solution.
Following similar approaches adopted in models relying on \svi, we devise a two-step optimization procedure for training the model.
Initially, we fix the variance of the variational parameters to low values in order to enforce stability in the optimization procedure during the early iterations.
We also fix the noise variance at all layers for the same purpose.
The former mitigates the risk of remaining stuck at the variational prior, while the latter trick is helpful for preventing the noise variance from becoming excessively large.
After a pre-established number of steps, the aforementioned parameters are then freed and trained jointly with the rest.
Further details on the set-up used for the experimental evaluation are given in Section~\ref{sec:experiments}.

Adapting the training procedure for \mfdgp to work with mini-batches is straightforward as it simply involves rescaling the model fit component appearing in Equation~\ref{eqn:marg_lik}.
The only caveat is in finding an adequate balance between observations having different fidelities in the composition of each mini-batch.
Assuming limited high-fidelity observations, one can include these at every training step while sub-sampling the data observed with lower fidelity.

\subsection{Complexity}\label{sec:complexity}

If we assume that the only observations available belong to the highest fidelity level, the computational complexity of the model is $\mathcal{O}\left(SNM^2\left( D_\text{out,1} + \dots + D_\text{out,L} \right) \right)$, which reduces to $\mathcal{O}\left(SNM^2L\right)$ in the case of having a single output dimension.
However, since we expect the majority of observations to be at lower-fidelity layers, training \mfdgp will be faster than a regular \dgp.
Our implementation of \mfdgp builds upon the \gpflow~\citep{Matthews2017} code provided for the model presented by~\cite{Salimbeni2017}, exploiting automatic differentiation for optimization.

\subsection{Comparison to \nargp and \deepmf}

Reframing the discussion in Section~\ref{sec:related_work} in view of the presented contributions, \mfdgp primarily distinguishes itself from \nargp in how intermediate \gps are linked.
Assuming nested input structures and no observation noise at lower fidelities, \cite{Perdikaris2017} show that the optimized posterior over the model parameters at level $t$ is optimal even if the \gps are trained sequentially in isolation (this is in sharp contrast to the visualization of our model given in Figure~\ref{fig:mfdgp_3l}, where fidelity levels are no longer disjointed).
While such constraints enable simpler and faster training, they are overly restrictive in practice since such guarantees are difficult to enforce when sourcing multi-fidelity data.
Our model lifts these constraints by introducing a singular objective (Equation~\ref{eqn:marg_lik}) with respect to which the inducing points and kernel parameters at all layers are jointly optimized.
This poses alternative modeling challenges which we address by leveraging advances in the specification of \dgps.
While signposted as a useful extension in earlier work, practical use of \svi for multi-fidelity modeling is also novel to this paper.

The \deepmf model \citep{Raissi2016} bears less resemblance to our model.
Its name is derived from a deep deterministic transformation that is applied to the inputs, but the multi-fidelity component of the model is identical to \ar.
Incorporating similar input transformations in our model would be straightforward, but we do not explore this option further here.
\begin{figure}[t!]
	\begin{center}
		\begin{tabular}{c}
			\begin{tabular}{cc}
				\footnotesize{\textsc{linear-a}} & \footnotesize{\textsc{linear-b}}\vspace{-.5ex}\\
				\includegraphics[width=100pt]{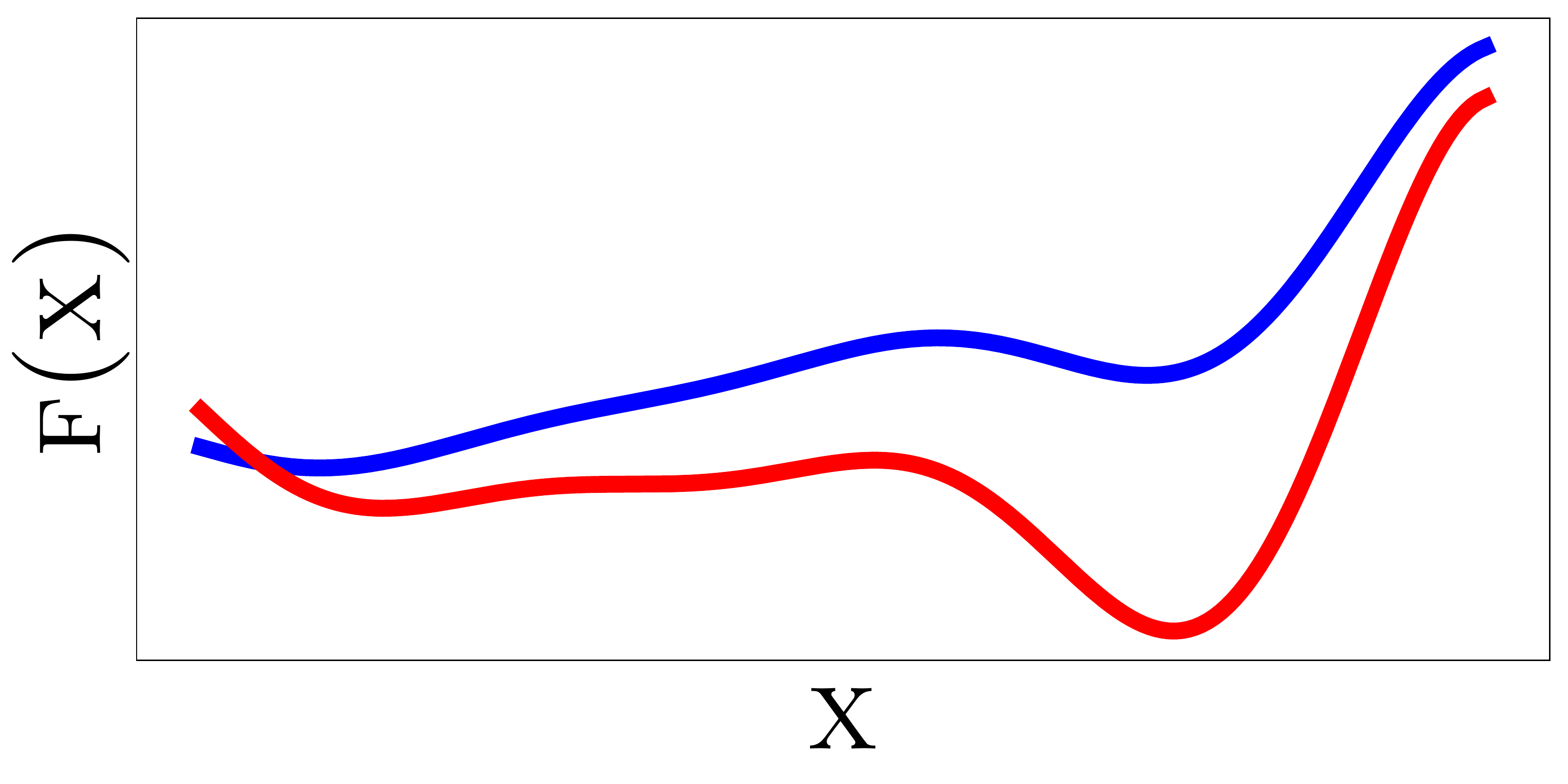} &
				\includegraphics[width=100pt]{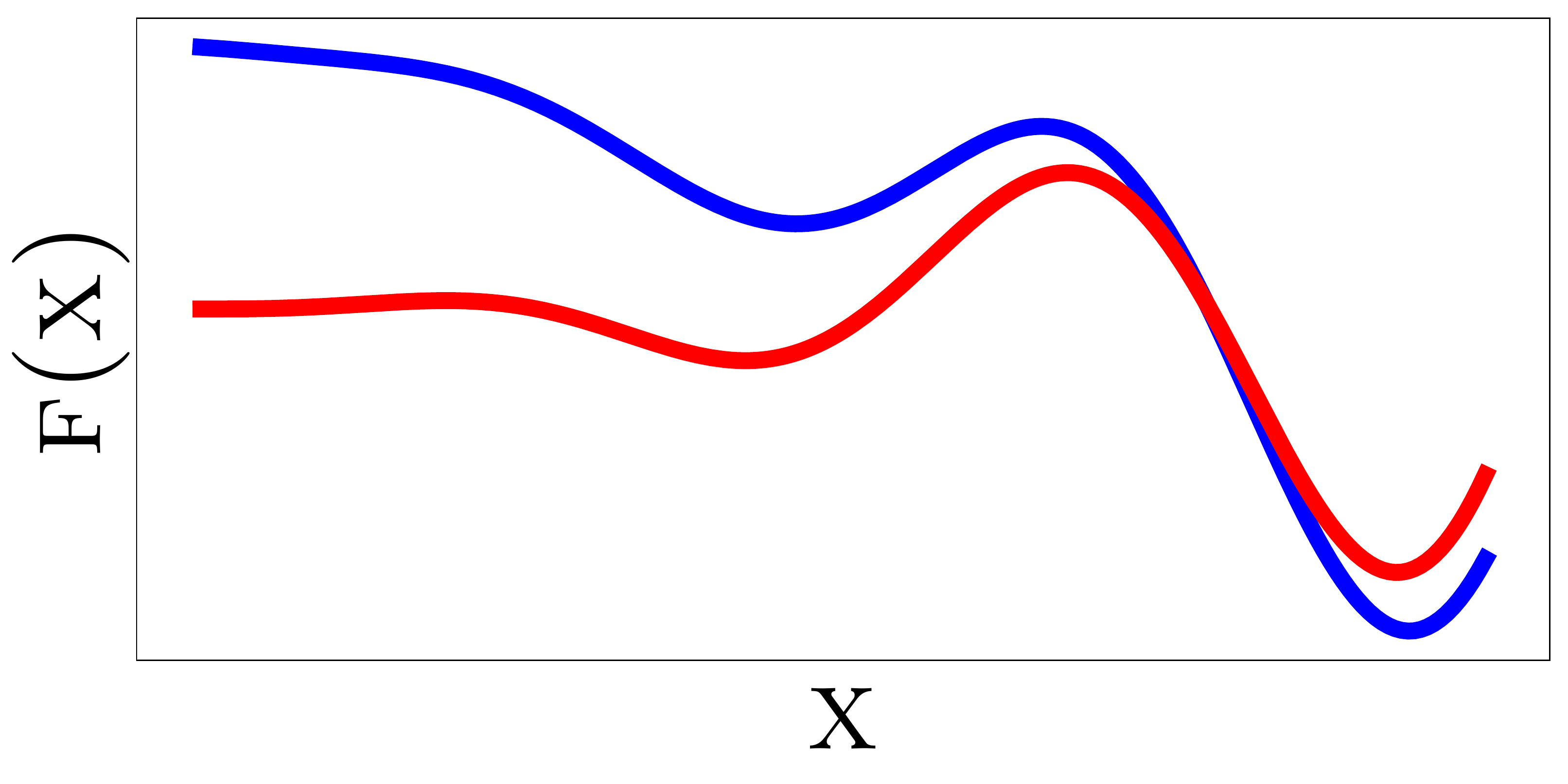}
			\end{tabular} \vspace{-1ex}\\
			\begin{tabular}{cc}
				\footnotesize{\textsc{nonlinear-a}} & \footnotesize{\textsc{nonlinear-b}}\vspace{-.5ex}\\
				\includegraphics[width=100pt]{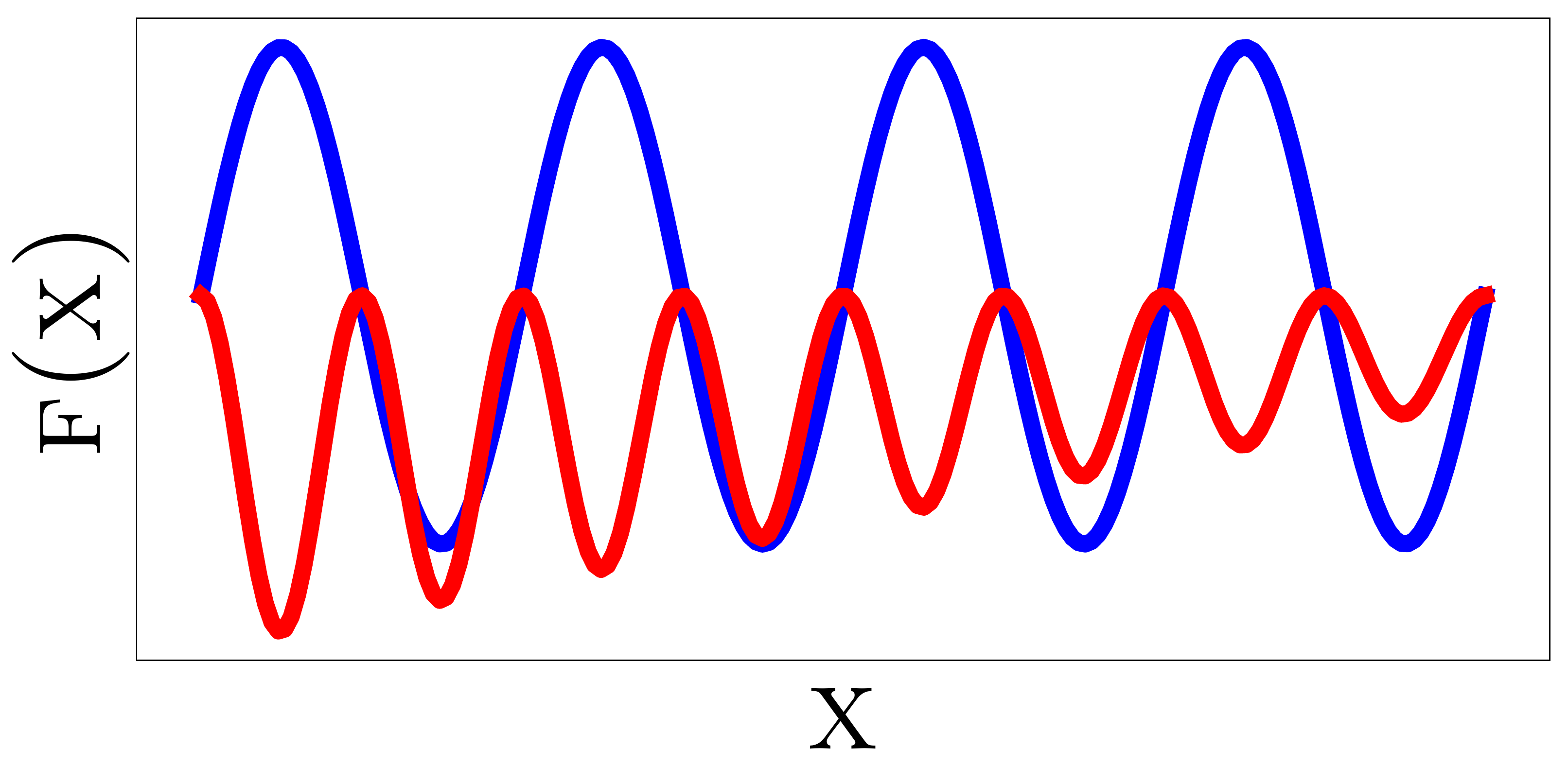} &
				\includegraphics[width=100pt]{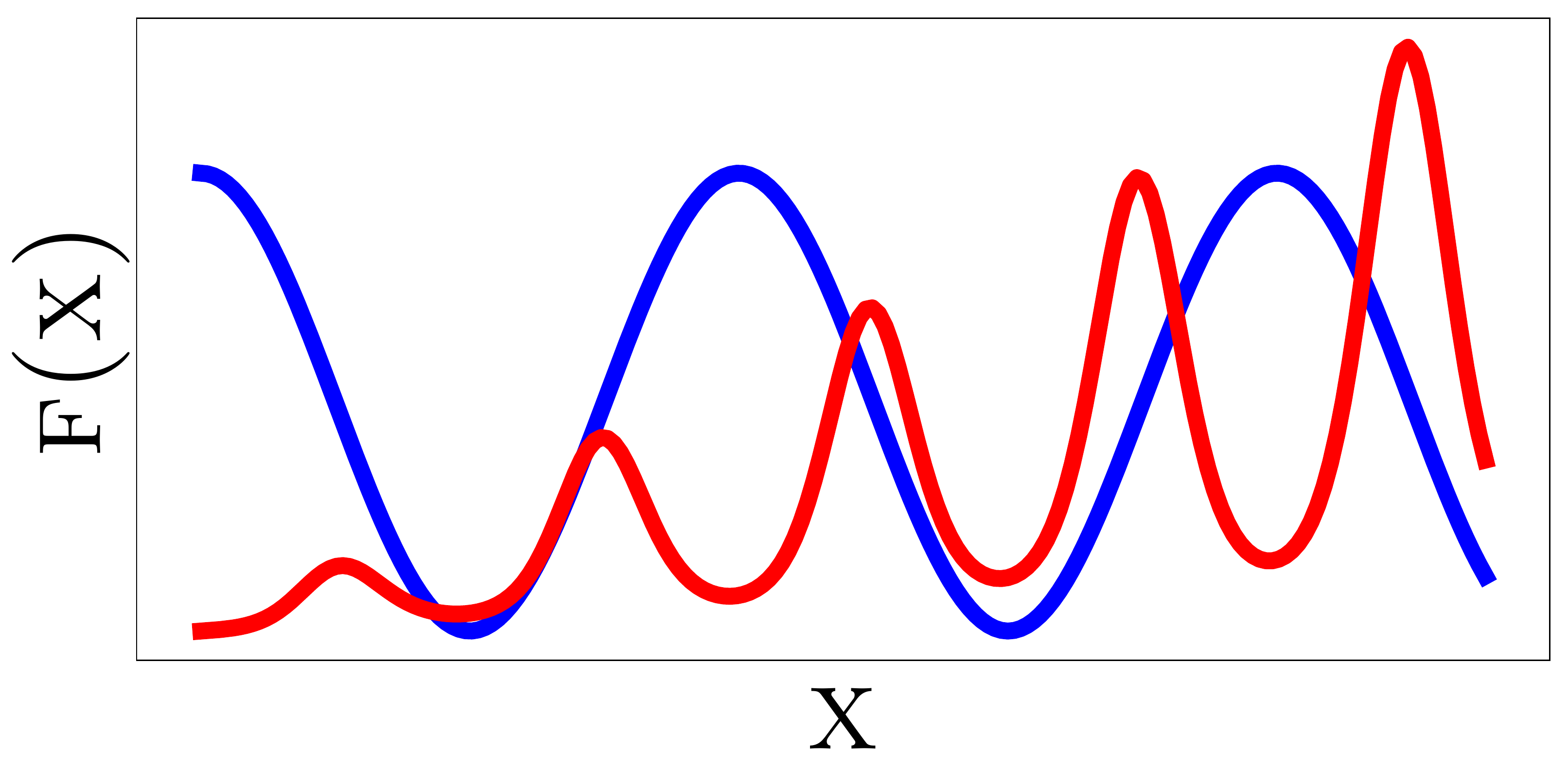} 
			\end{tabular}
			\vspace{-1ex}
		\end{tabular}
		\includegraphics[width=.3\textwidth]{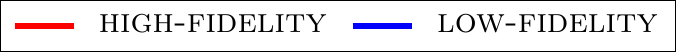}
	\end{center}
	\vspace{-1ex}
	\caption{Synthetic examples.
		\textit{Top: }Linear mapping between fidelities.
		\textit{Bottom: }Nonlinear mapping.}
	\label{fig:benchmarks_plot}
\end{figure}

\newcommand{\uqfigwidth}{95pt}
\begin{figure*}[t]
	\begin{center}
		\setlength{\tabcolsep}{4pt}
		\begin{tabular}{c@{\hspace{.3\tabcolsep}}cccc}
			&\textsc{\footnotesize{linear-a}}
			&\textsc{\footnotesize{linear-b}} &\textsc{\footnotesize{nonlinear-a}} &\textsc{\footnotesize{nonlinear-b}\vspace{-.05cm}}\\
			
			\multirow{1}{*}[5.5ex]{\parbox{2cm}{\footnotesize{\ar}}}&
			\includegraphics[width=\uqfigwidth]{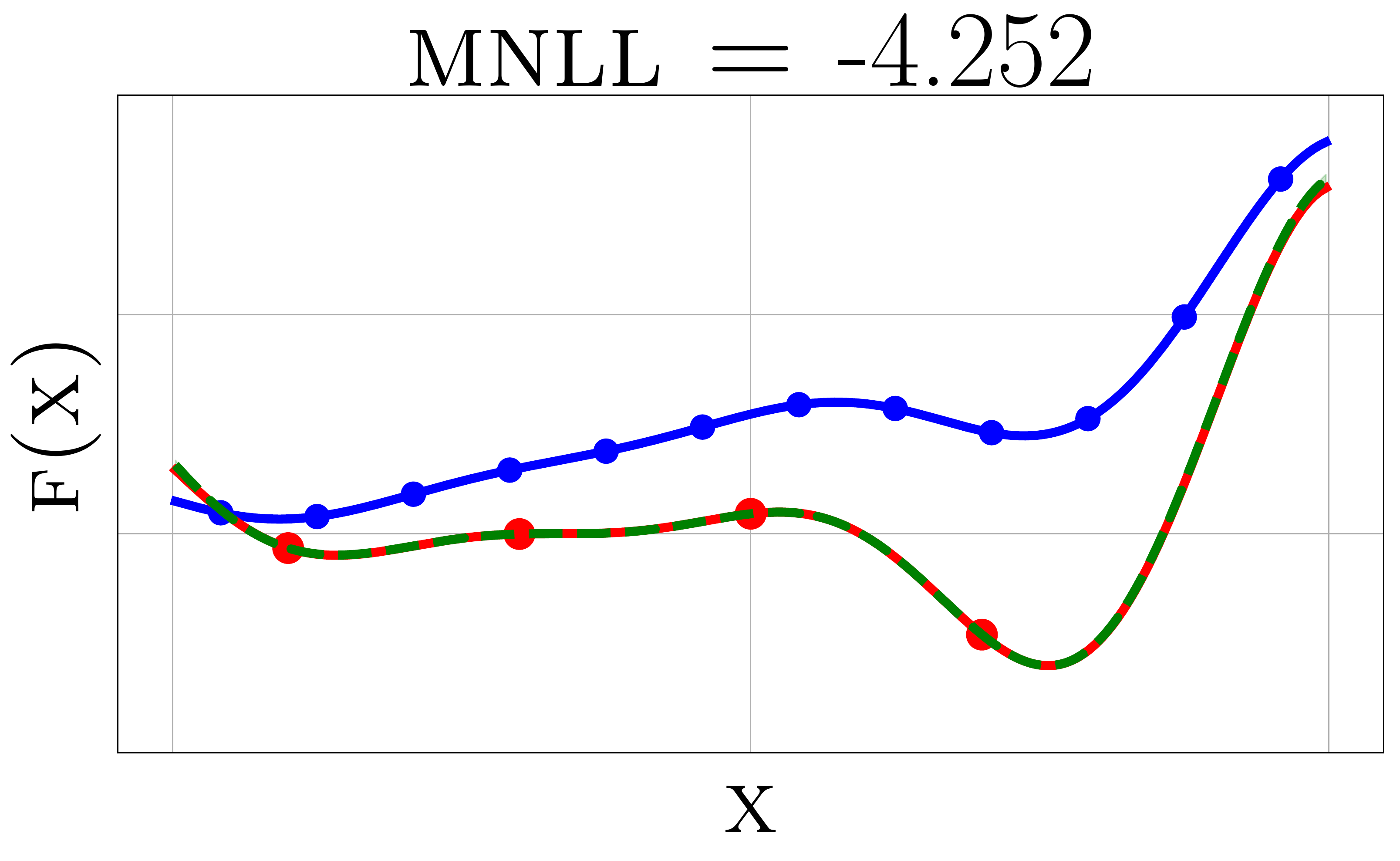} &
			\includegraphics[width=\uqfigwidth]{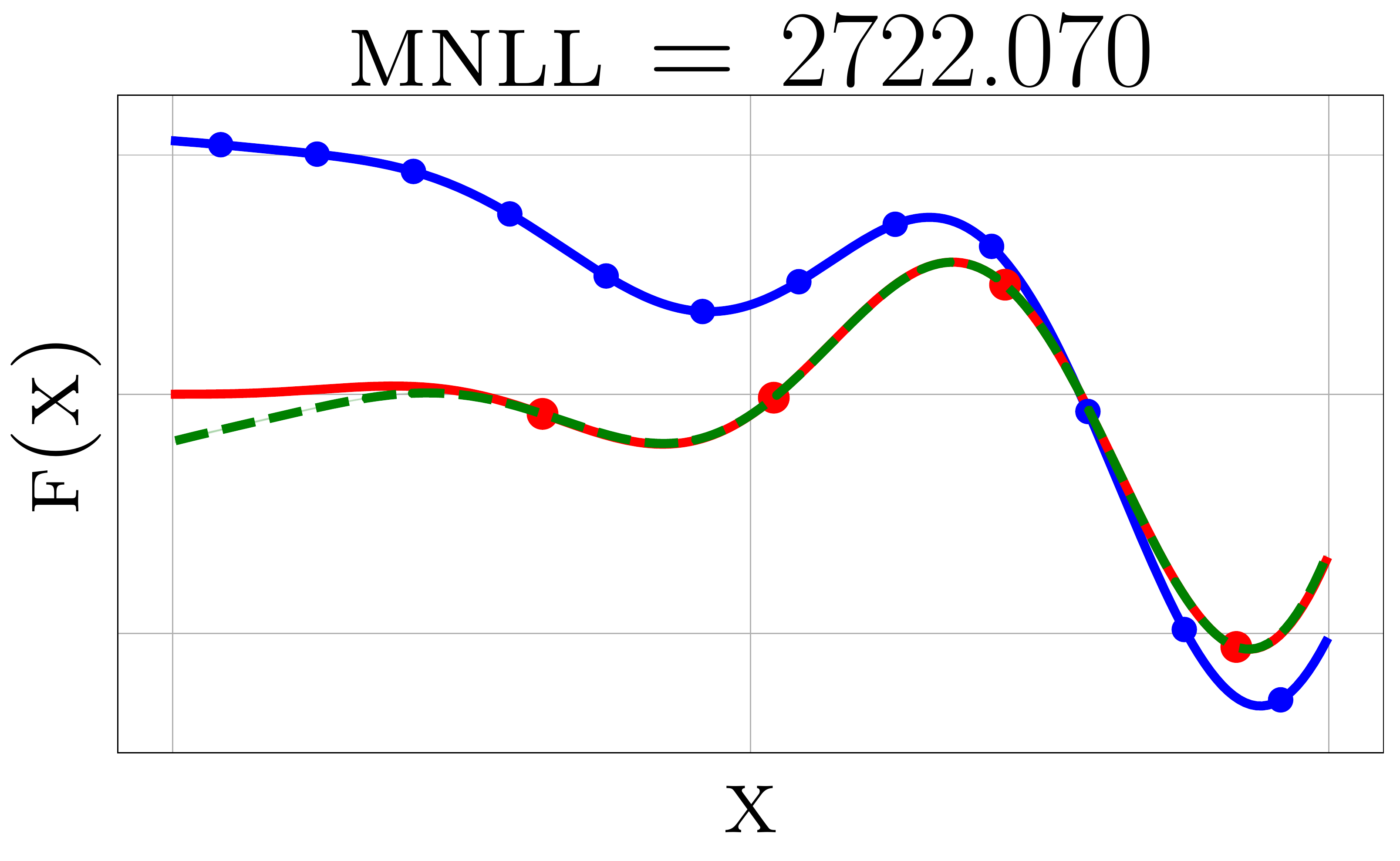} &
			\includegraphics[width=\uqfigwidth]{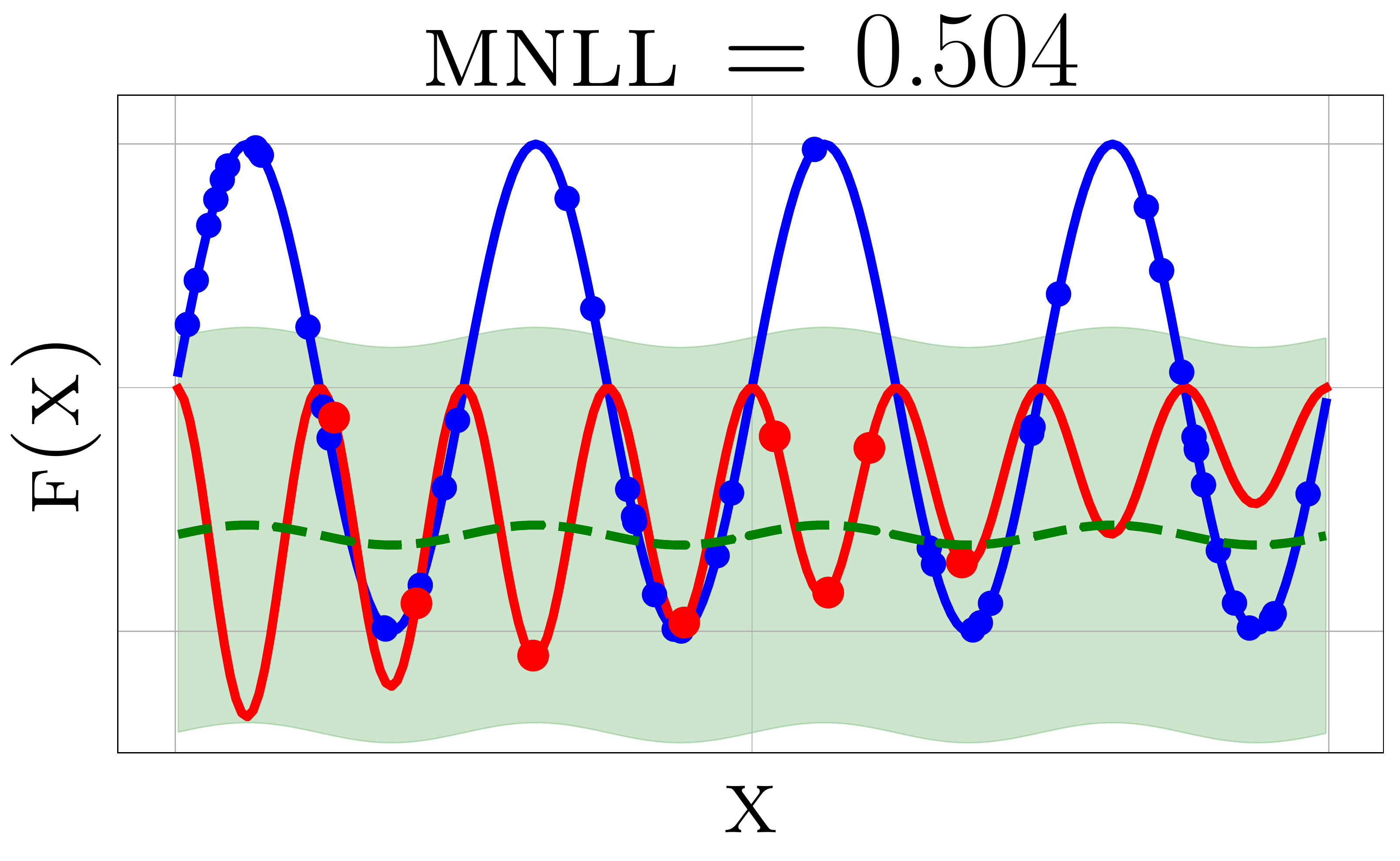} &
			\includegraphics[width=\uqfigwidth]{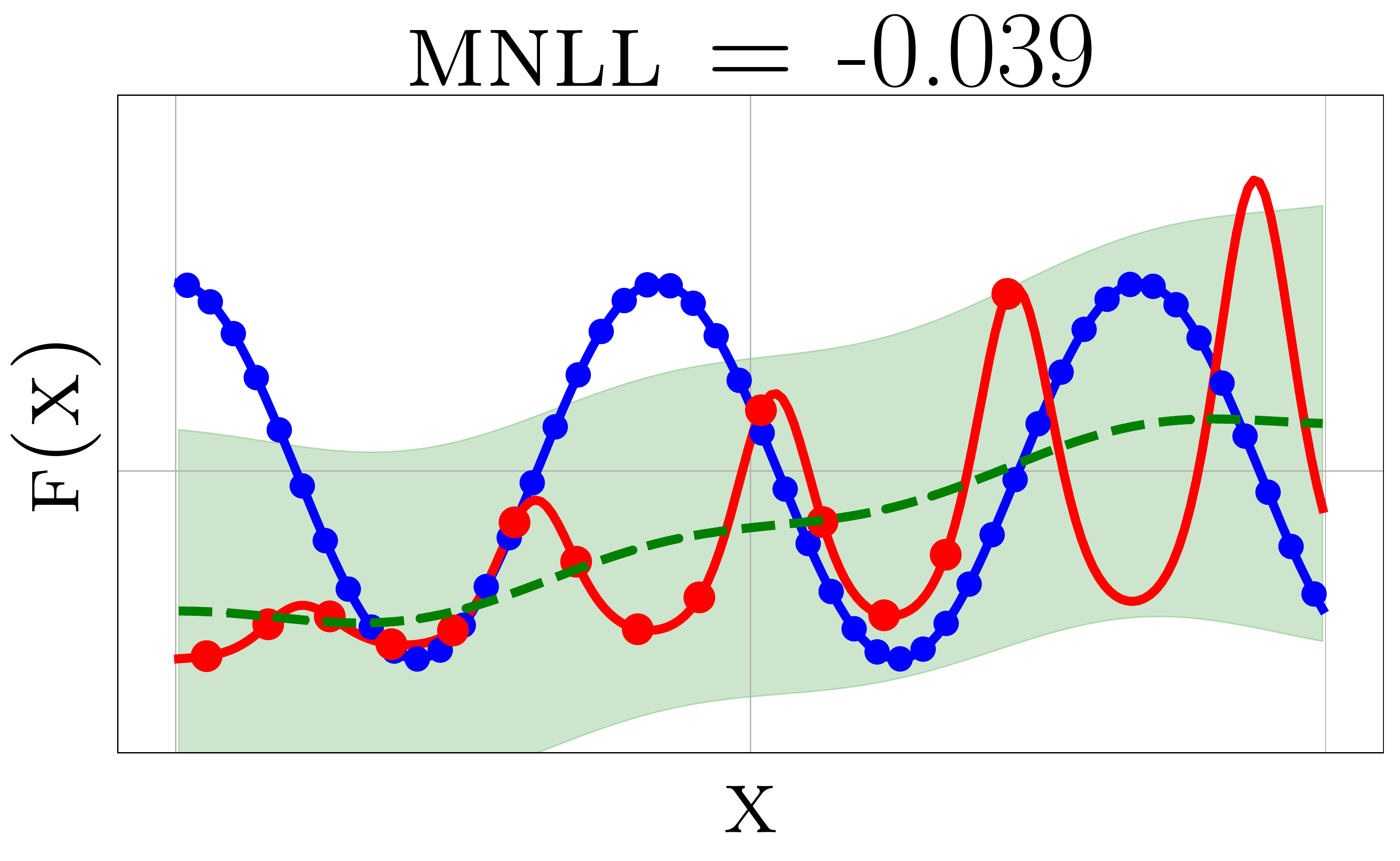}\vspace{-.1cm}\\
			
			\multirow{1}{*}[5.5ex]{\parbox{2cm}{\footnotesize{\nargp}}}&
			\includegraphics[width=\uqfigwidth]{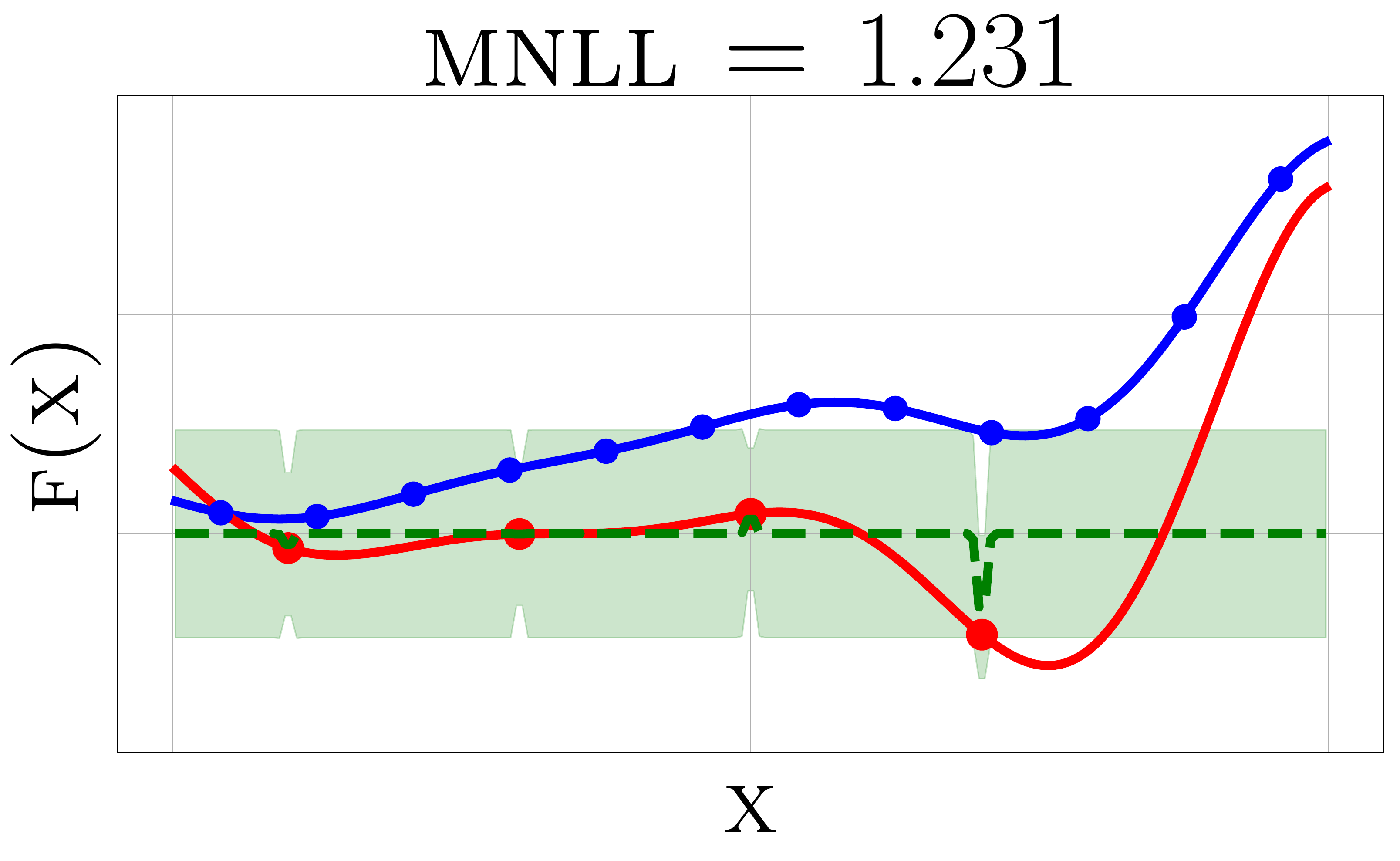} &
			\includegraphics[width=\uqfigwidth]{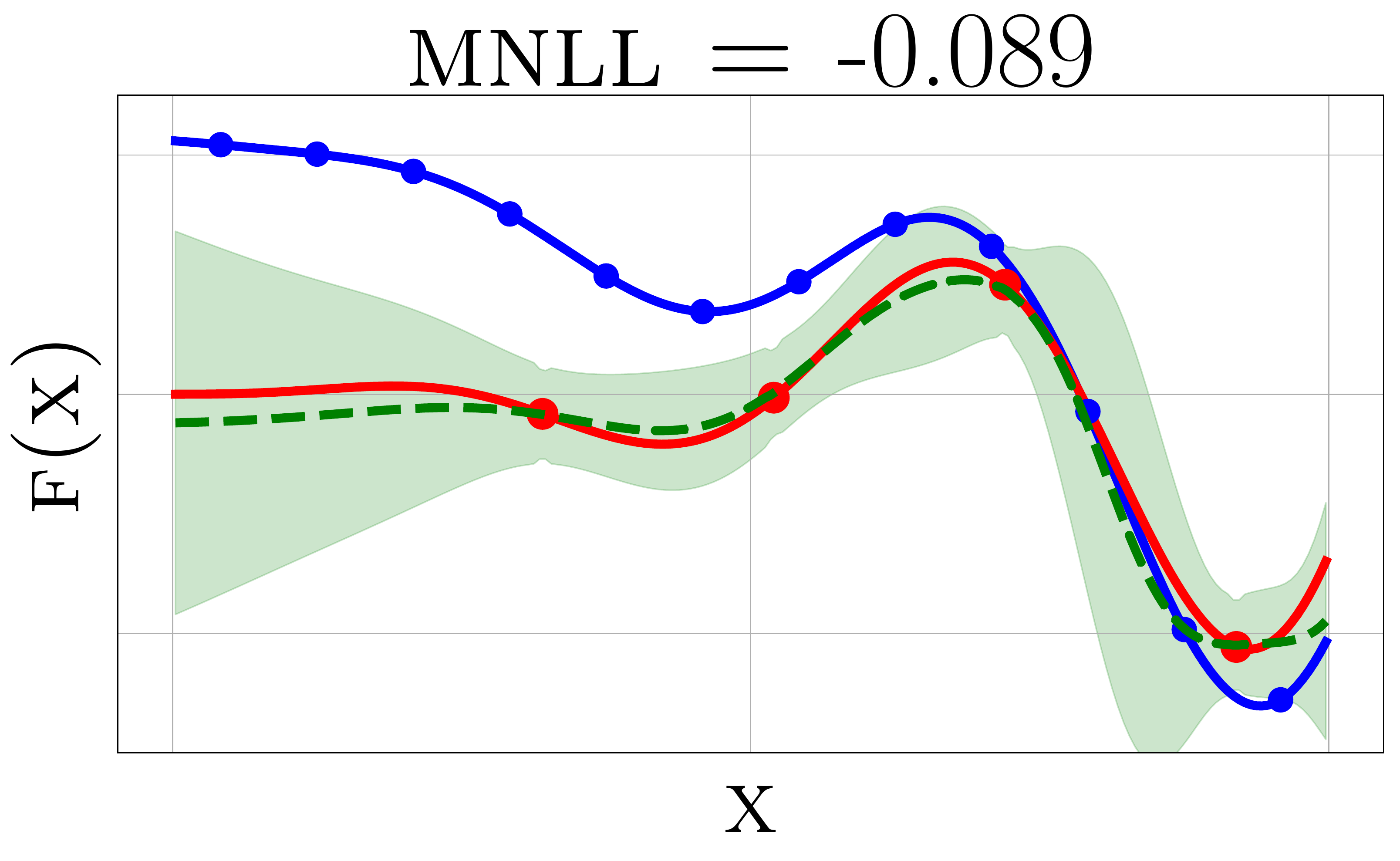} &
			\includegraphics[width=\uqfigwidth]{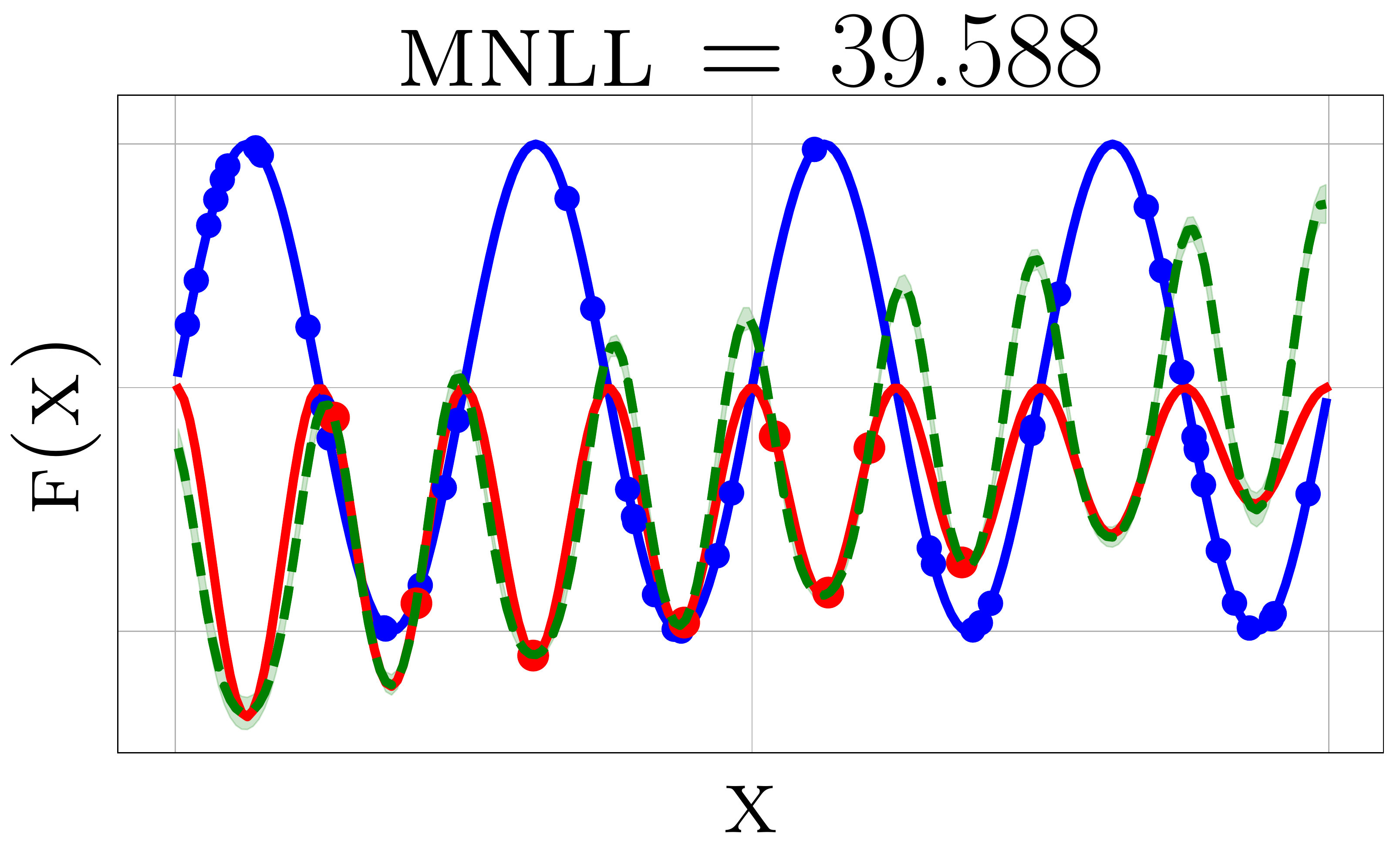} &
			\includegraphics[width=\uqfigwidth]{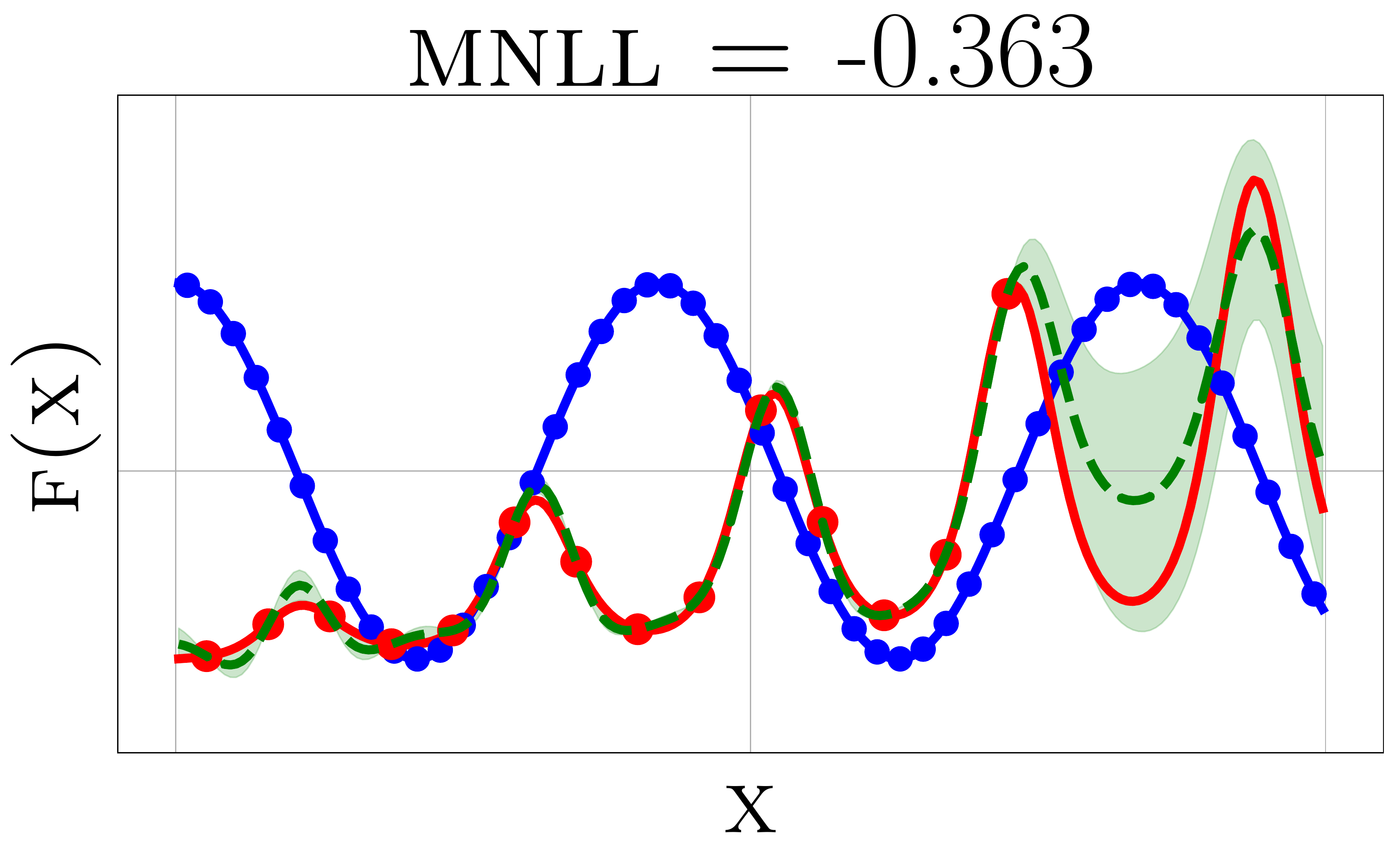}\vspace{-.1cm}\\
			
			\multirow{1}{*}[6.5ex]{\parbox{2cm}{\footnotesize{\dmf}}}&
			\includegraphics[width=\uqfigwidth]{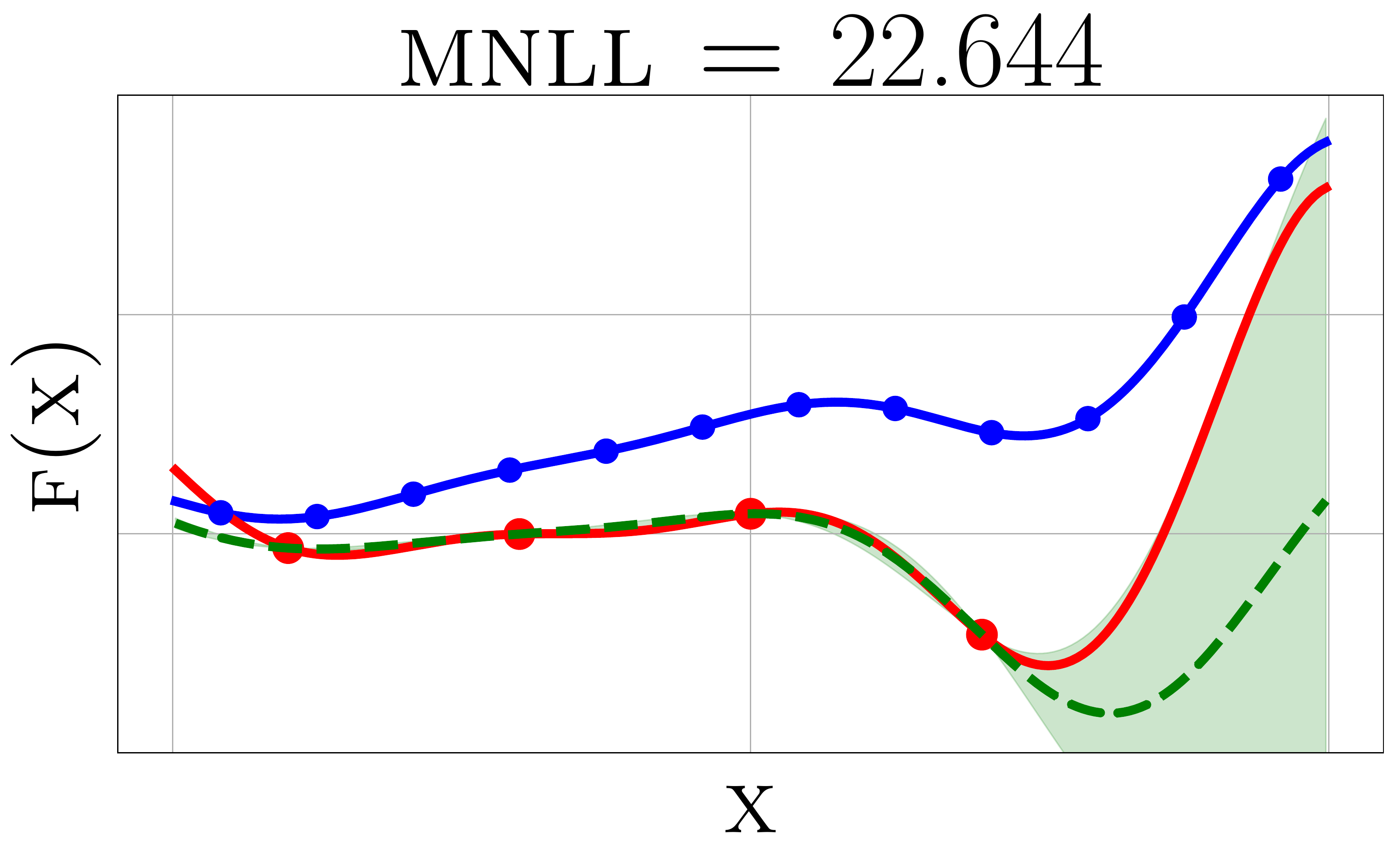} &
			\includegraphics[width=\uqfigwidth]{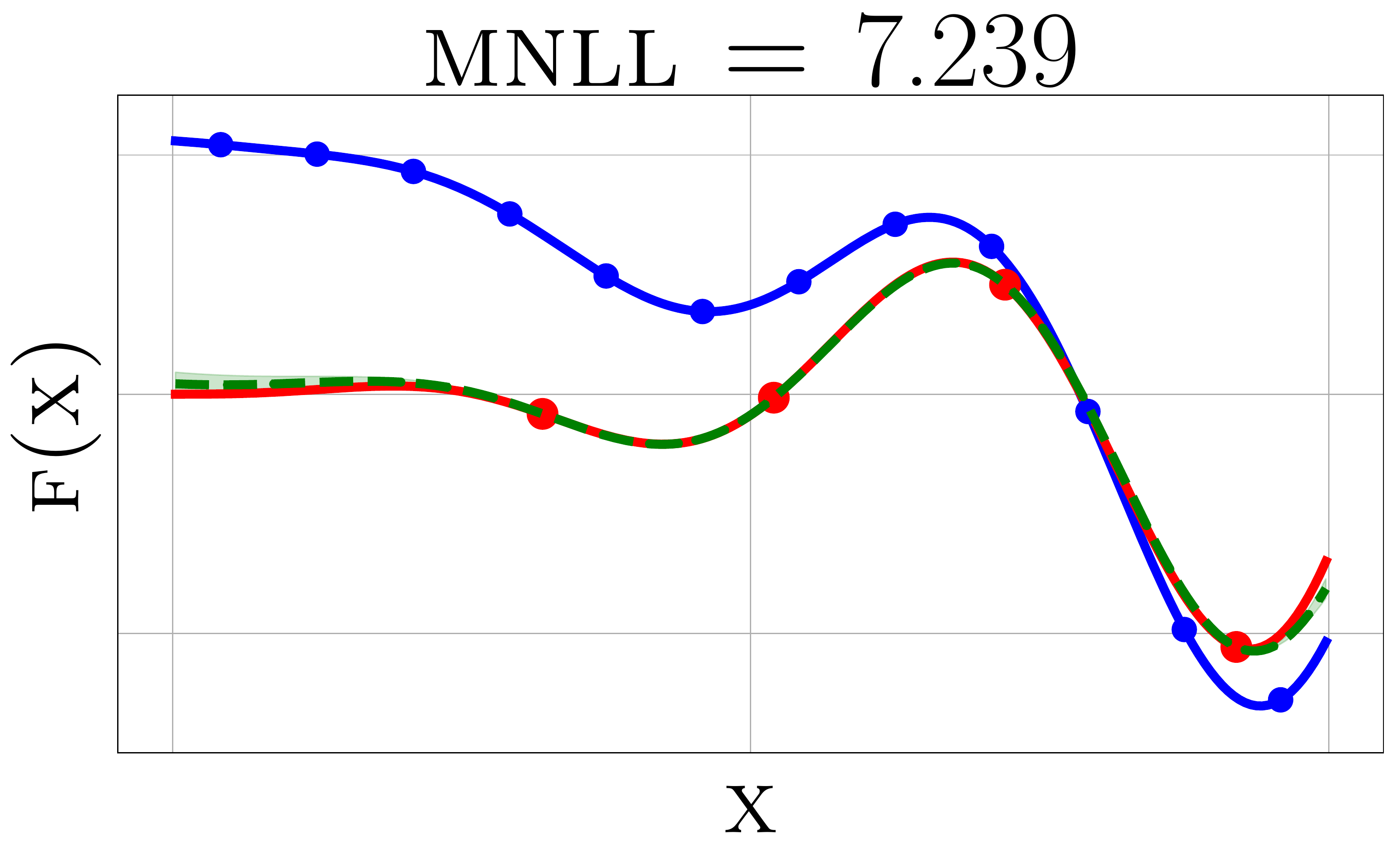} &
			\includegraphics[width=\uqfigwidth]{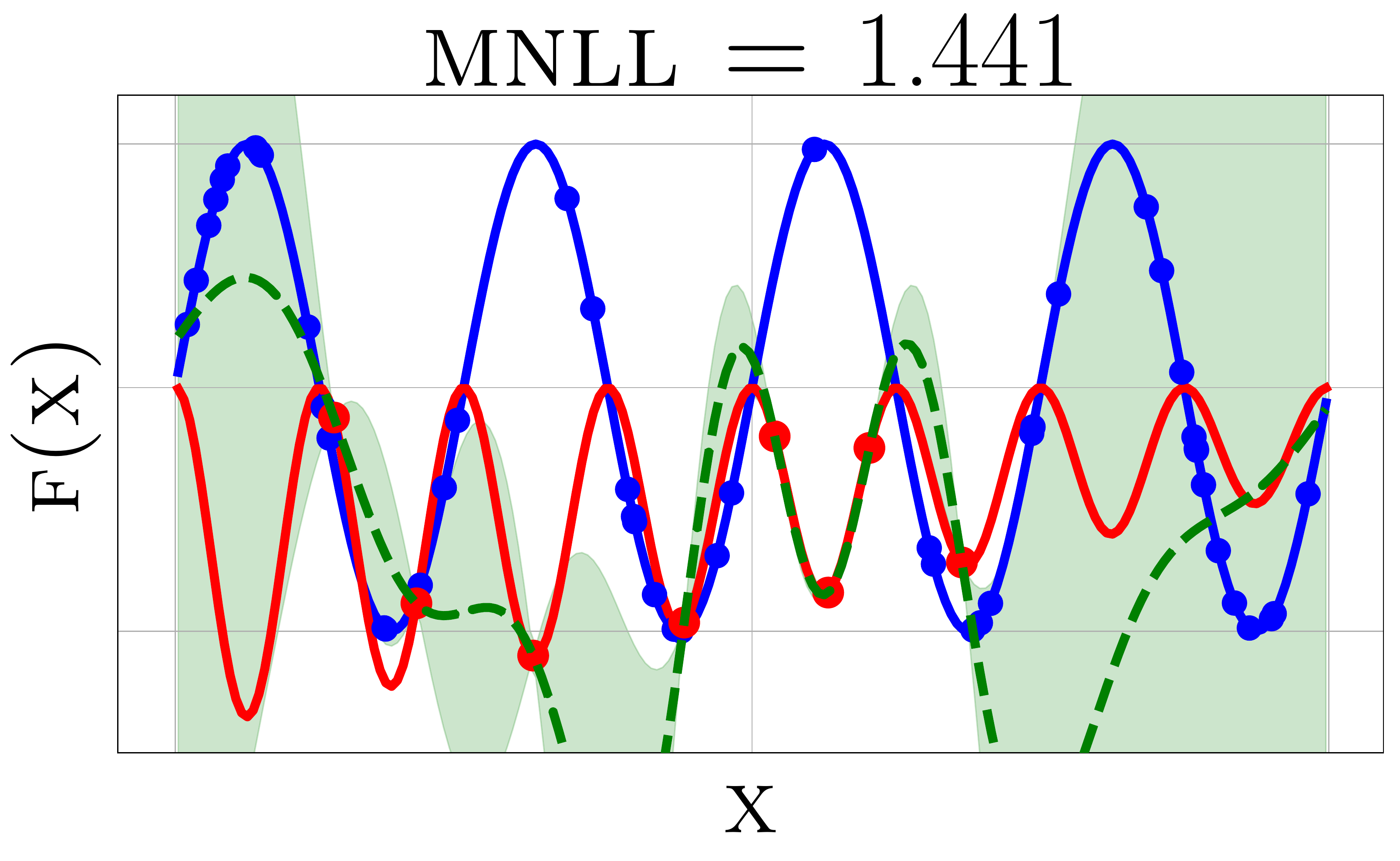} &
			\includegraphics[width=\uqfigwidth]{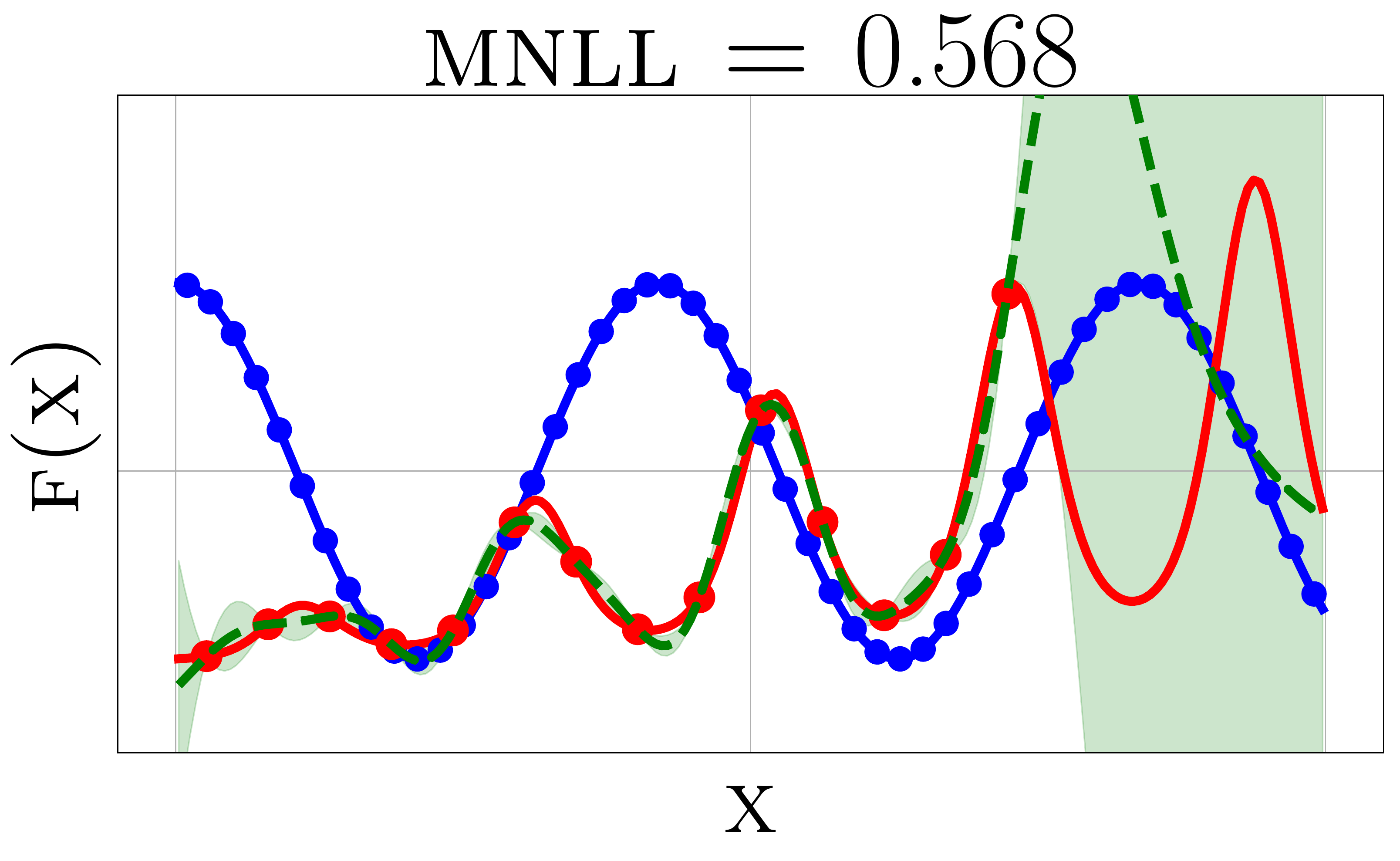}\vspace{-.1cm}\\
			
			\multirow{1}{*}[6.5ex]{\parbox{2cm}{\footnotesize{\mfdgp}}}&
			\includegraphics[width=\uqfigwidth]{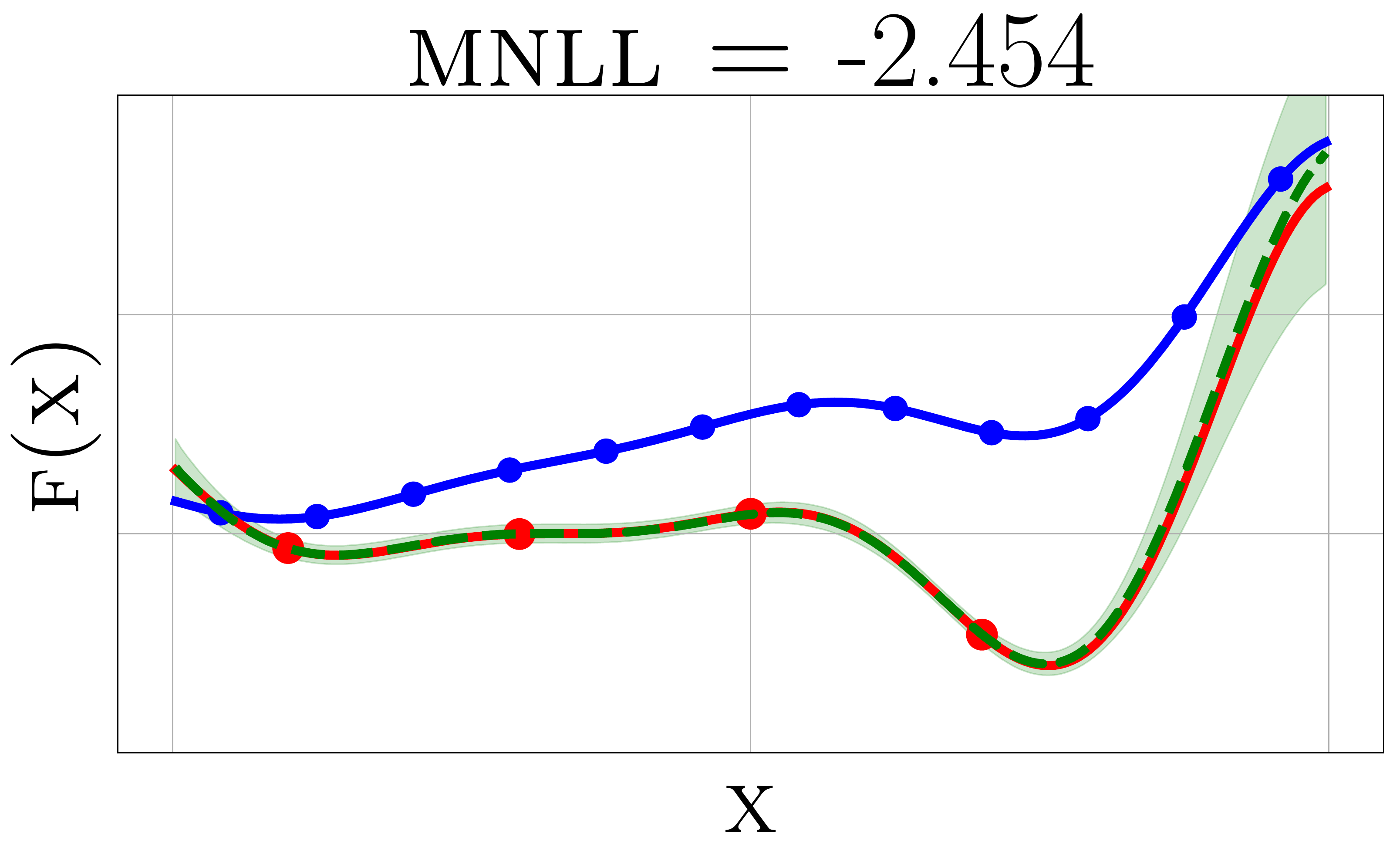} &
			\includegraphics[width=\uqfigwidth]{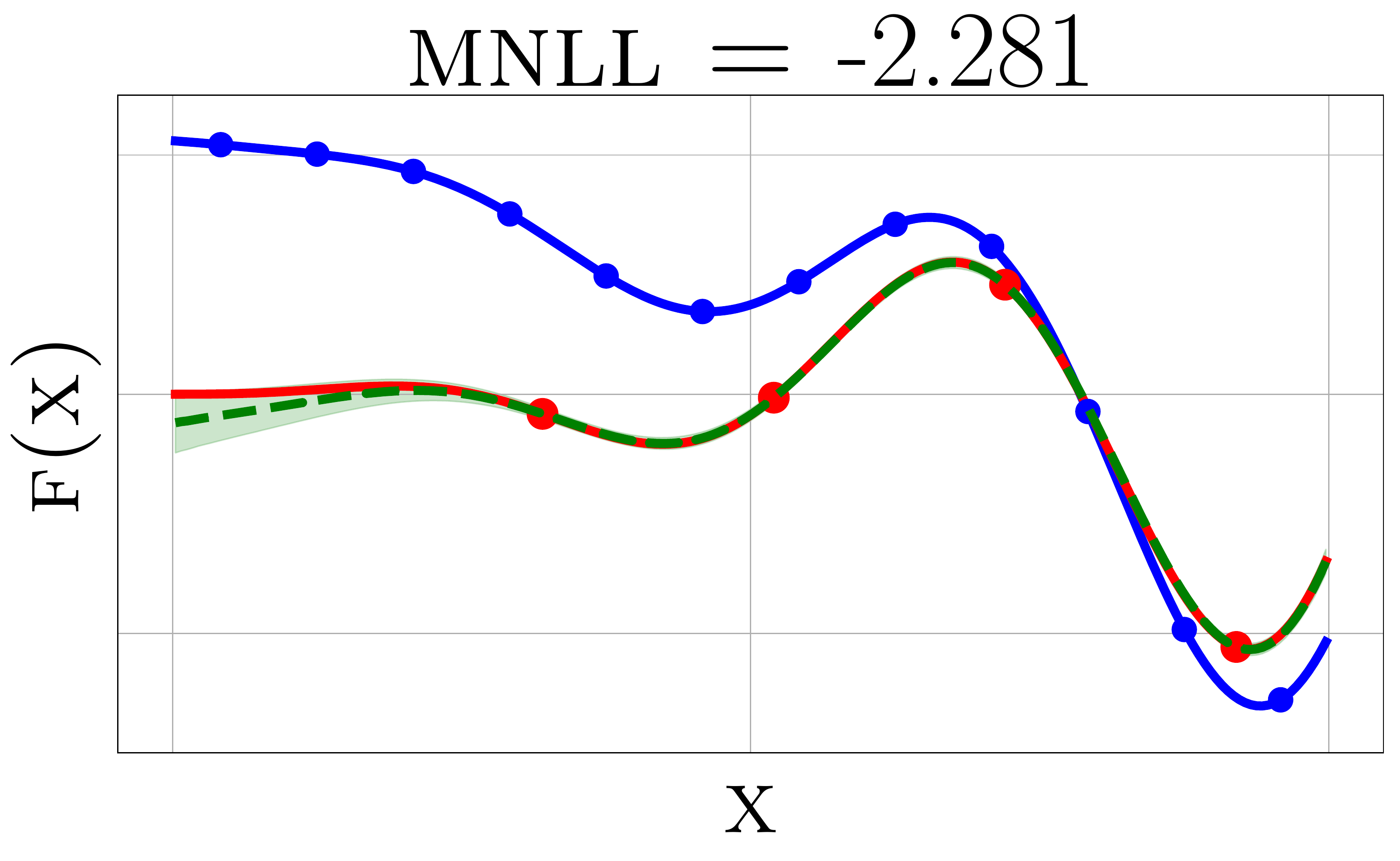} &
			\includegraphics[width=\uqfigwidth]{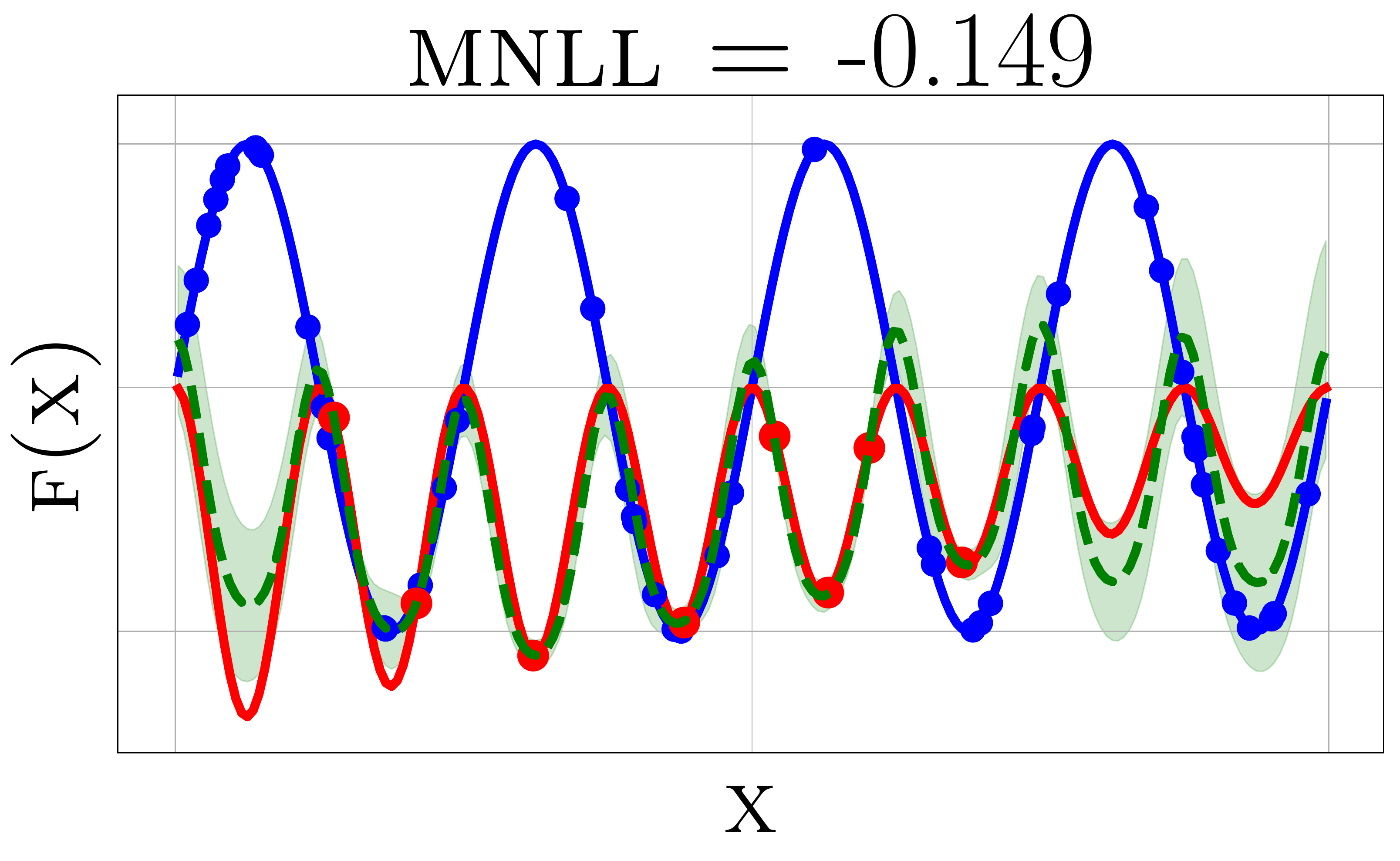} &
			\includegraphics[width=\uqfigwidth]{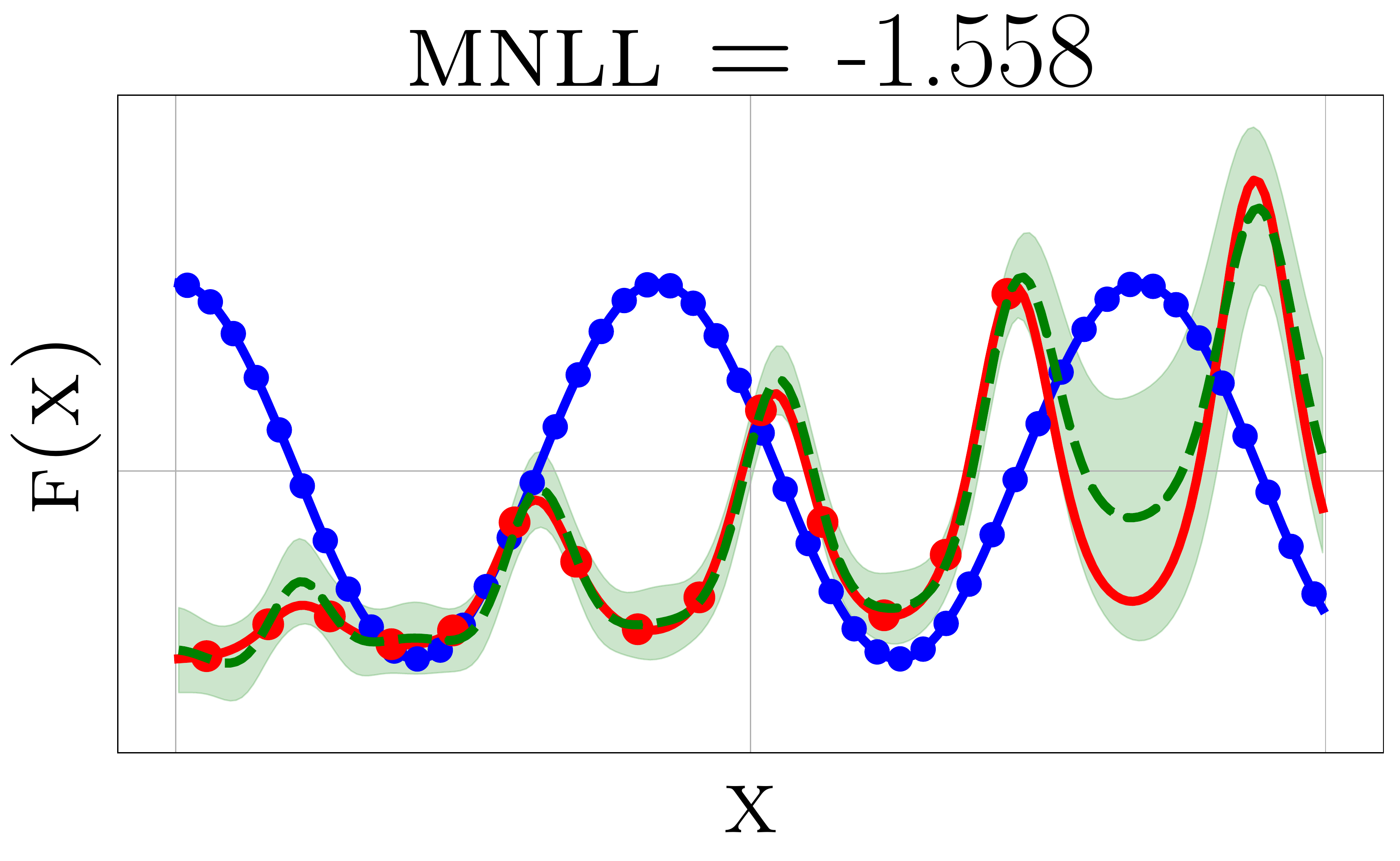}
		\end{tabular}
	\end{center}
		\vspace*{-5mm}
		\caption{Cross-comparison across methods and synthetic examples for challenging multi-fidelity scenarios.
		\mfdgp yields conservative uncertainty estimates where few high-fidelity observations are available.
	}
	\label{fig:uq_comparison}
\end{figure*}

\section{Experiments}\label{sec:experiments}

In the preceding sections, we developed a multi-fidelity model that can be trained end-to-end across fidelities.
Through a series of experiments, we validate that beyond its theoretic appeal, the proposed \mfdgp model also works well in practice.
We begin with a visual illustration of the superior uncertainty quantification returned by the model, and corroborate these findings by comparing it against competing techniques on a suite of established multi-fidelity problems with varying fidelity levels.
This is followed by an experiment involving a large-scale real-world dataset for which nearly a million observations are available.
An experimental design set-up showcasing the benefits of using \mfdgp in conjunction with determinantal point processes concludes this section.

\subsection{Synthetic Examples}

One of the primary motivations for undertaking this work was to develop a fully-fledged multi-fidelity model which avoids the overfitting issues encountered in existing models.
We commence this section by considering experimental set-ups where the available data is generally insufficient to yield confident predictions, and higher uncertainty is prized.
In particular, we consider four synthetic examples (plotted in Figure \ref{fig:benchmarks_plot}) - two where the correlation between fidelities is linear, and two where this is nonlinear.\footnote{Illustrations are given in the supplement.}
We train \mfdgp using the two-step procedure described in Section~\ref{sec:svi}, whereby the noise variance and variational parameters are fixed for the first 5,000 training steps, before being trained jointly with the rest for another 15,000 steps.
For increased stability, the variational distributions at lower fidelities are initially fixed to the known training targets; these are then freed and optimization is continued.
The Adam optimizer~\citep{Kingma2015} is used with learning rate set to 0.003 and 0.001 for the first and second training phases respectively.
Training generally converges in fewer iterations, but we keep this configuration for conformity.

In Figure \ref{fig:uq_comparison}, we compare our model to \ar, \nargp, and \deepmf on multi-fidelity scenarios where the allocation of high-fidelity data is either limited or constrained to one area of the input domain.
In all examples, our model yields appropriately conservative estimates in regions where insufficient observations are available.
The improved uncertainty quantification can be validated visually for these one-dimensional examples, but this is also corroborated by the superior mean negative log likelihood (\mnll) reported for \mfdgp on the test data.

\subsection{Benchmark Comparison}\label{sec:exp_benchmarks}

\begin{table*}[tbp!]
	\setlength{\tabcolsep}{4.5pt}
	\newcommand\Tstrut{\rule{0pt}{3ex}}
	\caption{Model Comparison on Multi-fidelity Benchmark Examples.}
	\vspace{-1.5ex}
%	\vspace*{-2.5mm}
	\begin{center}
		\begin{tabular}{lcc | ccc | ccc | ccc}
			 \hlineB{3}
			&&\textsc{fidelity} & \multicolumn{3}{c|}{\textsc{ar{\small 1}}} &\multicolumn{3}{c|}{\nargp} & \multicolumn{3}{c}{\mfdgp} \\
			\textsc{benchmark}& $D_\text{in}$  &\textsc{allocation} & \textsc{r$^2$} & \textsc{rmse} & \textsc{mnll} & \textsc{r$^2$} & \textsc{rmse} & \textsc{mnll} & \textsc{r$^2$} & \textsc{rmse} & \textsc{mnll} \\
			\hline
			\textsc{currin} & 2  & 12-5 & 0.913 & 0.677 & 20.105 & 0.903 & 0.740 & 20.817 & \textbf{0.935} & \textbf{0.601} & \textbf{0.763}  \Tstrut\\
			\textsc{park} &  4      & 30-5 & \textbf{0.985} & 0.575 & 465.377  & 0.954 & 0.928 & 743.119 & \textbf{0.985} & \textbf{0.565} & \textbf{1.383}\\
			\textsc{borehole} &8 & 60-5 & \textbf{1.000} & \textbf{0.005} & \textbf{-3.946} & 0.973 & 0.063 & -1.054 &  0.999 & 0.015 & -2.031\\
			\textsc{branin} &2 &80-30-10 & 0.891 & 0.044 &  -1.740 & 0.929 & 0.053 & -1.223 & \textbf{0.965 }& \textbf{0.030 }& \textbf{-2.572}\\
			\textsc{hartmann-{\small 3}d} &3 & 80-40-20 & \textbf{0.998} & \textbf{0.043} & 0.440 & 0.305 & 0.755 & 0.637 & 0.994 & 0.075 & \textbf{-0.731}\\
			\hline
		\end{tabular}
	\end{center}
	\label{table:benchmarks}
\end{table*}

\begin{figure*}[tbh!]
	\centering
	\begin{tabular}{ccc}
%		{\textsc{\footnotesize true}} & \footnotesize{\textsc{predicted}}& \footnotesize{\dpp}\vspace{-1.2ex}\\
		{\textsc{\footnotesize true high-fidelity}} & \footnotesize{\textsc{ predicted high-fidelity}}&\footnotesize{\textsc{\dpp samples}}\\
		\includegraphics[width=145pt]{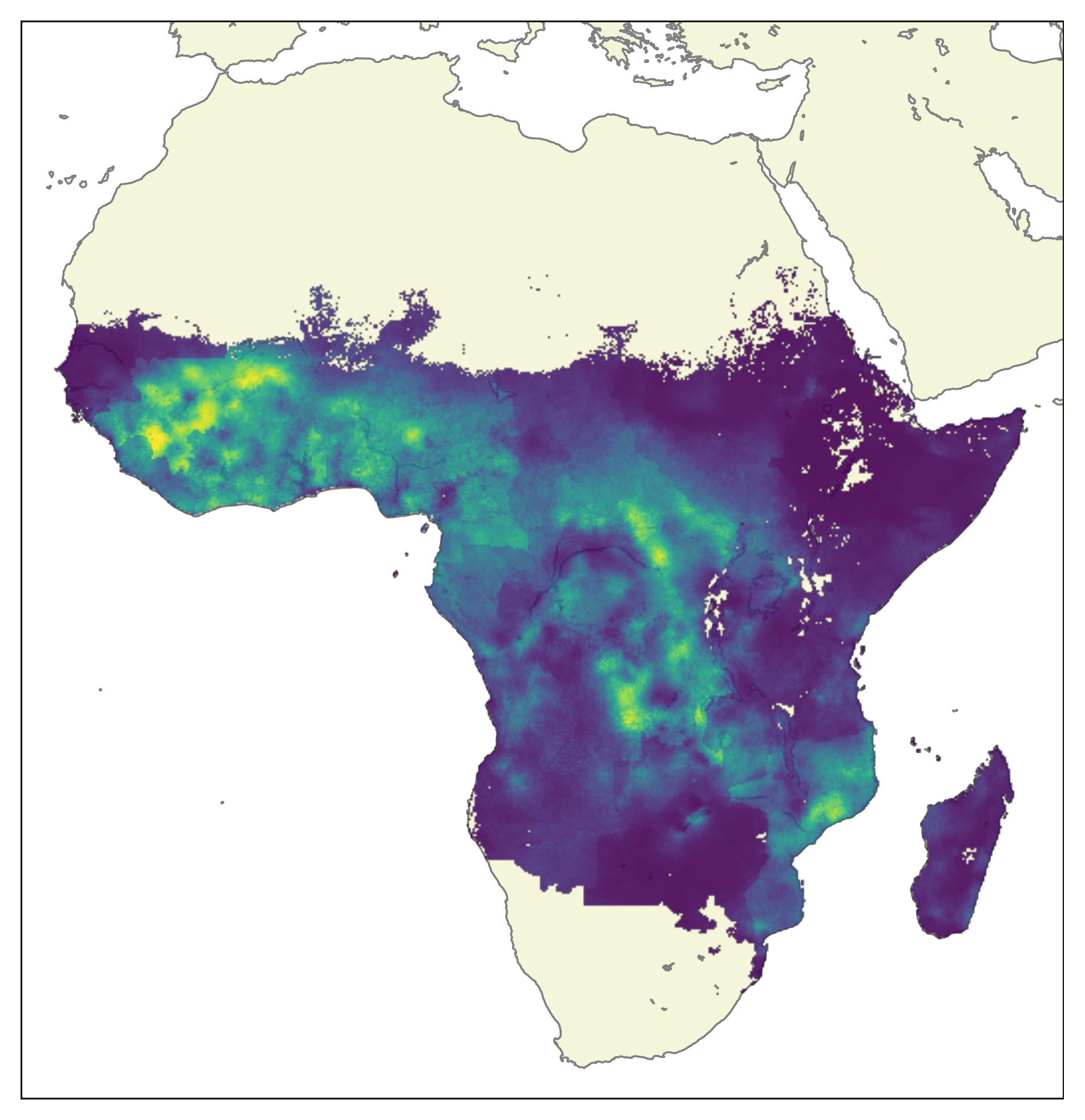} &
		\includegraphics[width=145pt]{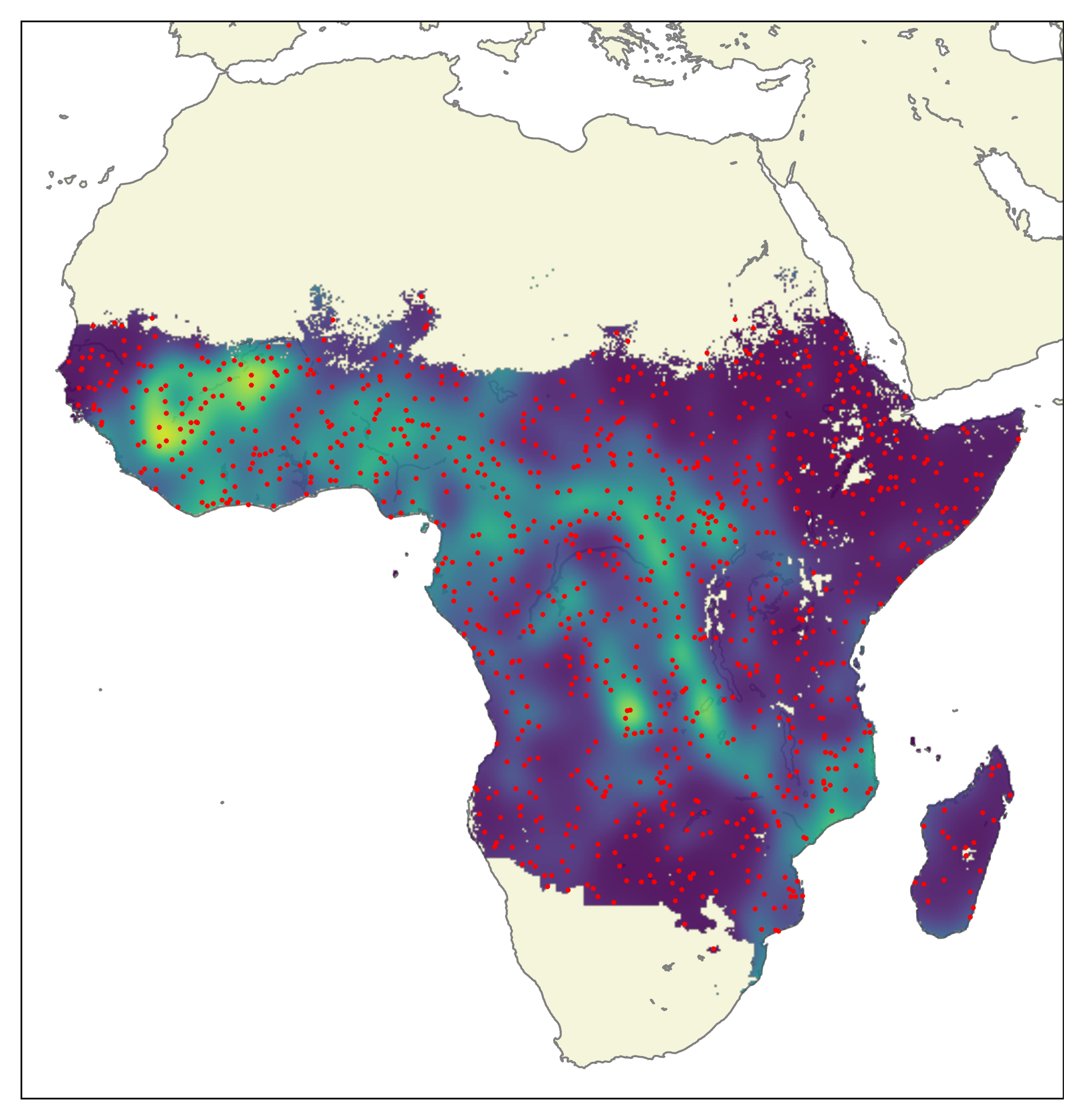} &
		\includegraphics[width=145pt]{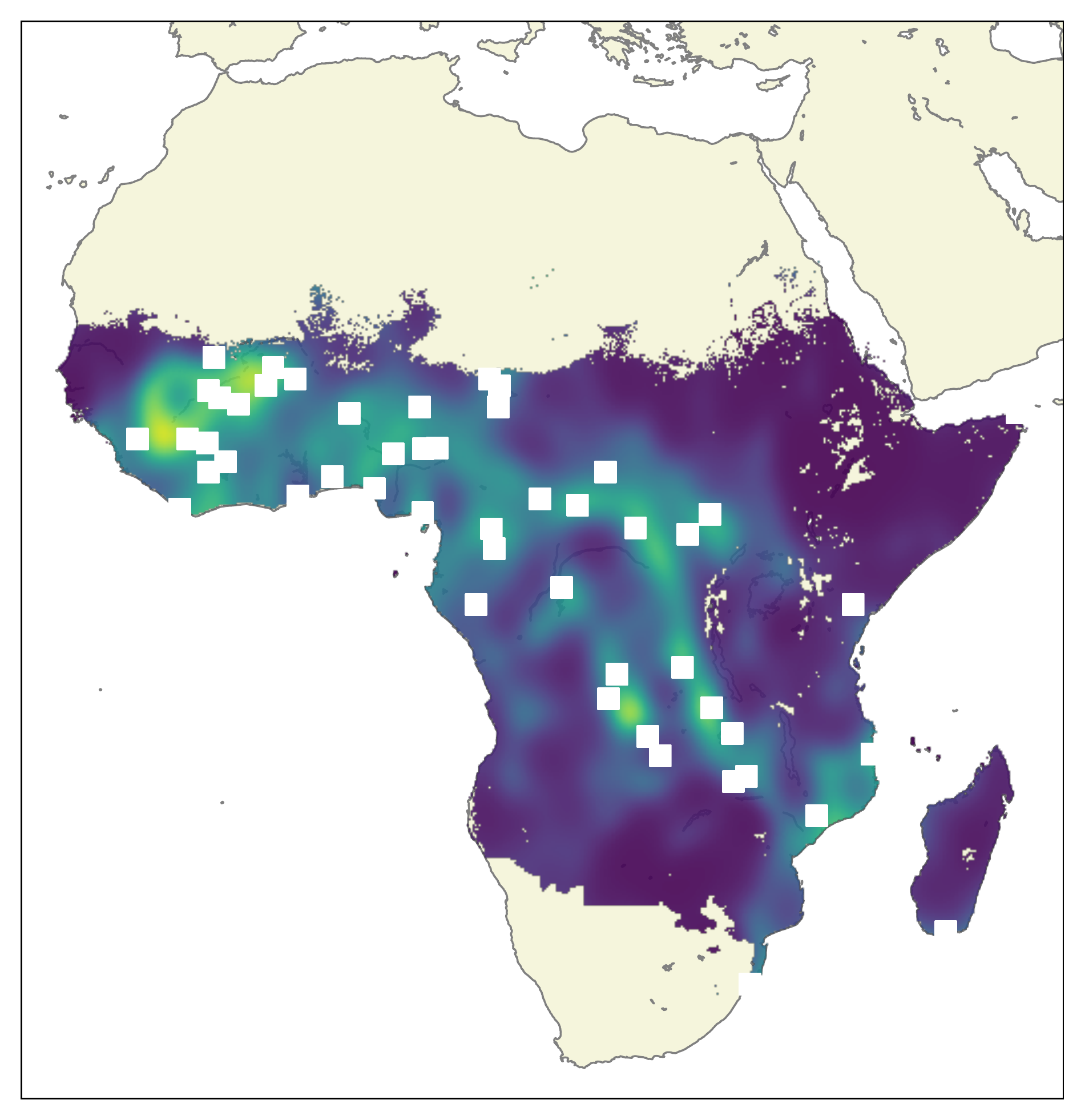}
	\end{tabular}
	\vspace*{-3mm}
	\caption{Real-world experiment indicating the infection rate of \textit{Plasmodium falciparum} among African children.
		Lighter-shaded regions denote higher infection rates in that area of the continent.
		\textit{Left: }True infection rates recorded for the year 2015.
		\textit{Center: }\mfdgp predictions given low-fidelity data from 2005 and limited high-fidelity training points (marked in red) from 2015. 
		\textit{Right: }White squares show the samples drawn from a \dpp using the posterior covariance of the \mfdgp model as its kernel.
	}
	\label{fig:malaria_plot}
\end{figure*}

Beyond the synthetically-constructed examples considered thus far, we verify the suitability of using \mfdgp over existing methods by benchmarking their performance on a selection of well-known multi-fidelity problems (full specification in the supplementary material).
Five randomly-generated datasets are prepared for each example function, following the allocation of points to different fidelities listed in Table~\ref{table:benchmarks}.
The results denote the $\textsc{R}$-squared $(\textsc{R}^2)$, root mean squared error (\textsc{rmse}), and \textsc{mnll} obtained using each model over a fixed test set of 1,000 points covering the entire input domain.
The obtained results give credence to our intuition that \mfdgp balances issues in alternative modeling approaches, which are singularly tailored for linear and nonlinear fidelity correlations respectively.
Notably, for the 3-level Branin function having nonlinear correlations between fidelities, the \ar model is incapable of properly modeling the high-fidelity data, whereas \mfdgp significantly outperforms \nargp on all metrics.

%%\subsection{Multi-fidelity in the Loop}

%%While deriving a multi-fidelity model with good predictive performance is already a noteworthy contribution in itself, the superior uncertainty estimates obtained using \mfdgp are salient in decision-making pipelines where well-calibrated uncertainty is essential.
%%We asses this quality using an expository experimental design loop where points are sequentially chosen to reduce uncertainty about a function of interest. 
%%In this experimental set-up we learn the borehole function by sequentially evaluating the function at high-fidelity points where the current variance of the high-fidelity predictive distribution is a maximum. [Results to come].

\subsection{Large-scale Real-world Experiment}

We now proceed to demonstrate the effectiveness of \mfdgp on a real-world dataset which also shows how mini-batch-based training with \svi is essential for modeling large datasets beyond the scale to which multi-fidelity methods are usually applied.
In particular, we fit \mfdgp to data describing the infection rate of \emph{Plasmodium falciparum} (a known cause of malaria) among children in Africa\footnote{Extracted from maps provided by The Malaria Atlas Project, \href{https://map.ox.ac.uk}{\text{https://map.ox.ac.uk}}.}, illustrated in Figure~\ref{fig:malaria_plot} (\textit{left}).
For our evaluation, we treat data from 2005 as being low-fidelity and more recent data from 2015 as high-fidelity; this permits us to exploit ample historical data to build an accurate model of the current infection rate for which fewer observations are given.
As the targets lie on the interval $[0, 1]$, we transform these using a logit function before fitting the model.

We train the model with 800,000 low-fidelity data-points and 1,000 high-fidelity observations, where each mini-batch consists of 1,000 low-fidelity and all 1,000 high-fidelity points.
Optimization is carried out using Adam for 30,000 iterations, while 1,000 inducing points are used at each layer.
Upon training, the model was evaluated on a test set comprising of 10,000 high-fidelity points.
The results obtained by \mfdgp on this data are visualized in Figure~\ref{fig:malaria_plot} (\textit{center}), with an \rmse of 0.063.
In contrast, an exact \gp trained only on high-fidelity observations scores an inferior \rmse of 0.096.

\subsection{Experimental Design with \mfdgp}

Lastly, we demonstrate how the posterior distribution associated with our \mfdgp model can be used for the purpose of experimental design.
In particular, we validate how this can be exploited in order to make decisions on where to obtain new observations of infection rates such that the overall quality of predictions returned by the model is improved.
We are generally interested in observing these new points with high-fidelity at locations where either uncertainty is large (leading to a more diverse set of locations) or where we expect there to be a substantial infection rate (denoted by lighter shading on the map).
This balances the exploration-exploitation trade-off that is commonly targeted by such schemes.

A determinantal point process \citep[\dpp;][]{Macchi1975} is well-suited for addressing the aforementioned criteria; the kernel function of a \dpp is chosen to be $\mu(\mathbf{x})k(\mathbf{x}, \mathbf{x}')\mu(\mathbf{x}')$, where $k(\cdot, \cdot)$ and $\mu(\cdot)$ denote the posterior covariance and mean functions of the \mfdgp model. 
The covariance term encourages points to be selected at a set of diverse locations where the model uncertainty is high, whereas the mean term gives greater weight to input locations where the infection rate is expected to be high. 
In order to sample from the \dpp, we first evaluate the mean and covariance of the trained \mfdgp at a randomly-selected set of 2,500 input locations.
By setting the cardinality $k=50$, a $k$-\dpp \citep{Kulesza2011} is then used to sample 50 high-fidelity points from this subset, which are then interpreted as the locations at which true infection rates should be acquired.
Extending the experiment presented in the previous section, the sampled points are illustrated by white markers in Figure \ref{fig:malaria_plot} (\textit{right}).
Recalling the criteria highlighted at the beginning of this section, the plot clearly indicates that the points selected by this procedure are adequately dispersed across the map, with increased concentration in areas where infection rates are predicted to be high.
This validates the suitability of our multi-fidelity model in a pipelined decision-making scheme.
\section{Conclusion}\label{sec:conclusion}

Reliable decision making under uncertainty is a core requirement in multi-fidelity scenarios where unbiased observations are scarce or difficult to obtain.
In this paper, we proposed the first complete specification of a multi-fidelity model as a \dgp that is capable of capturing nonlinear relationships between fidelities with reduced overfitting.
By providing end-to-end training across all fidelity levels, \mfdgp consistently yields superior quantification and propagation of uncertainty that is crucial in active learning and iterative methods such as experimental design.
The application of state-of-the-art \dgps to an unconventional setting is also essential for broadening their appeal to a wider community of researchers and practitioners alike.

Effectively optimizing the inducing variables at each layer while remaining faithful to the implicit multi-fidelity constraints is a challenging problem which warrants further investigation, and is key to extending the learning capacity of \mfdgp.
On another note, in contrast to the standard \ar model, the compositional structure of \mfdgp hinders the specification of analytic expressions for the acquisition functions prevalent in procedures such as Bayesian optimization or quadrature.
Beyond the multi-fidelity setting explored here, the latter requirement accentuates ongoing effort to develop active learning schemes that are better-suited for deep models.

\bibliographystyle{icml2017}
\bibliography{references}

\appendix
\onecolumn

\title{Deep Gaussian Processes for Multi-fidelity Modeling}

\author{}

\section{Further Model Details}

For completeness, in the following appendix we extend the model description given in Section 4 of the main paper.
In particular, we detail the variational approximation for the model and derive the evidence lower bound that serves the role of our model's multi-fidelity objective function with respect to which parameters are optimized.
As in the main text, we intentionally remain faithful to the general notation and structure of~\cite{Salimbeni2017} in order to place emphasis on the multi-fidelity extension being proposed in this work as opposed to the \dgp approximation upon which it is based.

\subsection{Approximating the Marginal Likelihood of \mfdgp}

Assume that each layer, $l$, of our \mfdgp model corresponds to a realisation of the process modeled with fidelity $t$.
For a dataset with $T$ fidelities, the marginal likelihood of our \mfdgp model is then given by:

\begin{equation}\label{eqn:loss_mfdgp_3l}
	\mathcal{L_{\mfdgp}} = \mathbb{E}_{ q \left(\lbrace \{\fvect_l^t\}^t_{l=1}\rbrace_{t=1}^T, \lbrace \uvect_l\rbrace^L_{l = 1}\right)} \left[\log\left( \frac{p\left(\lbrace \textbf{y}^t, \{\fvect_l^t\}_{l=1}^t\rbrace_{t=1}^T, \lbrace\uvect_l\rbrace^L_{l = 1}\right) }
	{ q \left(\lbrace \{\fvect_l^t\}^t_{l=1}\rbrace_{t=1}^T , \lbrace \uvect_l\rbrace^L_{l = 1}\right)} \right)\right].
\end{equation}

We have that:
%\begin{align}\label{eqn:loss_mfdgp_num_3l}
%p\left(\lbrace \textbf{Y}^l, \{\mathbf{F}_l^t\}_{t=1}^L, \mathbf{U}_l\rbrace^L_{l = 1}\right) = 
%	& \prod_i^{P_L}p\left( \mathbf{y}^L_i | f^L_{i,L} \right) \prod_{l=1}^L p\left( F^l_L | \mathbf{U}^l; F^{l-1}_L, \mathbf{Z}^{l-1} \right) p\left( \mathbf{U}^L | \mathbf{Z}^{L-1} \right) \times \nonumber \\
%	& \prod_j^{P_{L-1}}p\left( \mathbf{y}^{L-1}_j | f^{L-1}_{j,{L-1}} \right) \prod_{l=1}^{L-1} p\left( F^l_{L-1} | \mathbf{U}^l; F^{l-1}_{L-1}, \mathbf{Z}^{l-1} \right) p\left( \mathbf{U}^{L-1} | \mathbf{Z}^{L-2} \right) \times \dots \times \nonumber \\
%	&  \prod_k^{P_1}p\left( \mathbf{y}^1_k | f^1_{k,1} \right) \prod_{l=1}^1 p\left( F^l_1 | \mathbf{U}^l; F^{l-1}_1, \mathbf{Z}^{l-1} \right) p\left( \mathbf{U}^1 | \mathbf{Z}^{0} \right)
%\end{align}
\begin{align}\label{eqn:loss_mfdgp_num_3l}
p\left(\lbrace \yvect^t, \{\fvect_l^t\}_{l=1}^t\rbrace_{t=1}^T, \lbrace\uvect_l\rbrace^L_{l = 1}\right) = 
&\prod_{t=1}^T\prod^{N_t}_{i=1} p\left( {y}^{i,t} | {f}^{i,t}_t \right) \prod_{l=1}^t p \left( \fvect^t_{l} | \uvect_l; \{\fvect^t_{l-1}, \textbf{X}^t\}, \textbf{Z}_{l-1} \right) \times \nonumber \\
&  \prod_{l=1}^{L} p\left( \uvect_l ; \mathbf{Z}_{l-1} \right),
\end{align}

where $N_t$ denotes the number of data points observed with fidelity level $t$. 
Similarly the denominator in the expectation can be expanded as:
%\begin{align}\label{eqn:loss_mfdgp_denom_3l}
%	q \left(\lbrace \{\mathbf{F}_l^t\}_{t=1}^L, \mathbf{U}_l\rbrace^L_{l = 1}\right) =  
%	& \prod_{l=1}^L p\left( F^l_L | \mathbf{U}_l; F^{l-1}_L, \mathbf{Z}_{l-1} \right) q\left( \mathbf{U}_L\right) \times \nonumber \\
%	& \prod_{l=1}^{L-1} p\left( F^l_{L-1} | \mathbf{U}^l; F^{l-1}_{L-1}, \mathbf{Z}^{l-1} \right) q\left( \mathbf{U}^{L-1}\right) \times \dots \times \nonumber \\
%	& \prod_{l=1}^1 p\left( F^l_1 | \mathbf{U}^l; F^{l-1}_1, \mathbf{Z}^{l-1} \right) q\left( \mathbf{U}^1\right) 
%\end{align}
\begin{align}\label{eqn:loss_mfdgp_denom_3l}
q \left(\lbrace \{\fvect_l^t\}^t_{l=1}\rbrace_{t=1}^T, \lbrace \uvect_l\rbrace^L_{l = 1}\right) =  
& \prod_{t=1}^T\prod_{l=1}^t p\left( \fvect_l^t | \uvect_l; \{\fvect^t_{l-1}, \textbf{X}^t\}, \mathbf{Z}_{l-1} \right) \times \nonumber \\
& \prod_{l=1}^L q\left( \uvect_l\right) .
\end{align}

By inserting Equations (\ref{eqn:loss_mfdgp_num_3l}) and (\ref{eqn:loss_mfdgp_denom_3l}) in (\ref{eqn:loss_mfdgp_3l}), and canceling out equivalent terms in the numerator and denominator, we obtain the following variational lower bound on the marginal likelihood of our multi-fidelity model:

\begin{align}\label{eqn:loss_exp_3l}
	\mathcal{L_{\mfdgp}} 
	&= \iint q \left(\lbrace \{\fvect_l^t\}^t_{l=1}\rbrace_{t=1}^T , \lbrace \uvect_l\rbrace^L_{l = 1}\right) \log \left(
	\frac{ \prod_{t=1}^T\prod^{N_t}_{i=1} p\left( {y}^{i,t} | {f}^{i,t}_t \right) \times \prod_{l=1}^{L} p\left( \uvect_l ; \mathbf{Z}_{l-1} \right) }
	{  \prod_{l=1}^L q\left( \uvect_l\right)  }
	\right) \nonumber\\
	&\quad\text{d} \lbrace \{\fvect_l^t\}^t_{l=1}\rbrace_{t=1}^T , \lbrace \uvect_l\rbrace^L_{l = 1} \nonumber \\
	&=  \iint q \left(\lbrace \{\fvect_l^t\}^t_{l=1}\rbrace_{t=1}^T, \lbrace \uvect_l\rbrace^L_{l = 1}\right) \log \left( \prod_{t=1}^T\prod^{N_t}_{i=1} p\left( {y}^{i,t} | {f}^{i,t}_t \right) \right) \text{d} \lbrace \{\fvect_l^t\}^t_{l=1}\rbrace_{t=1}^T , \lbrace \uvect_l\rbrace^L_{l = 1} \nonumber \\
	& \quad +  \iint  q \left(\lbrace \{\fvect_l^t\}^t_{l=1}\rbrace_{t=1}^T, \lbrace \uvect_l\rbrace^L_{l = 1}\right) \log \left( 
	\frac{ \prod_{l=1}^L p\left(\uvect_l; \mathbf{Z}_{l-1}\right)}
	{\prod_{l=1}^L q\left(\uvect_l\right)} 
	\right) \text{d} \lbrace \{\fvect_l^t\}^t_{l=1}\rbrace_{t=1}^T , \lbrace \uvect_l\rbrace^L_{l = 1} \nonumber \\
	&=  \int q \left( \lbrace \{\fvect_l^t\}^t_{l=1}\rbrace_{t=1}^T \right) \log \left( \prod_{t=1}^T\prod^{N_t}_{i=1} p\left( {y}^{i,t} | {f}^{i,t}_t \right) \right) \text{d} \lbrace \{\fvect_l^t\}^t_{l=1}\rbrace_{t=1}^T \nonumber \\
	& \quad +  \int q \left(\lbrace \uvect_l\rbrace^L_{l = 1}\right) \log \left( 
	\frac{ \prod_{l=1}^L p\left(\uvect_l; \mathbf{Z}_{l-1}\right)}
	{\prod_{l=1}^L q\left(\uvect_l\right)} 
	\right) \text{d} \lbrace \uvect_l\rbrace^L_{l = 1} \nonumber \\
	& =  \sum_{t=1}^{T} \int q \left(  \{\fvect_l^t\}^t_{l=1} \right) \log \left( \prod_{i=1}^{N_t}p\left( {y}^{i,t} | {f}^{i,t}_t \right)\right) \text{d} \{\fvect_l^t\}^t_{l=1}\nonumber \\
	& \quad + \sum_{l=1}^L D_\text{KL}\left[ q\left(\uvect_l\right) ||\;p\left(\uvect_l; \mathbf{Z}_{l-1}\right)\right] \nonumber\\
	&=  \sum_{t=1}^T\sum_{i=1}^{N_t}\mathbb{E}_{q\left({f}^{i,t}_{t}\right)}\left[ \log p\left({y}^{i,t} | {f}^{i,t}_t\right) \right] \nonumber \\
	& \quad + \sum_{l=1}^L D_\text{KL}\left[q\left(\uvect_l\right) ||\;p\left(\uvect_l; \mathbf{Z}_{l-1}\right)\right].
\end{align}

If both the true distribution and the variational approximation are assumed to be Gaussian, the $D_\text{KL}$ term can conveniently be evaluated analytically.

\subsection{Reparameterization Trick}

As with other \dgp models~\citep{Dai2016, Cutajar2017} trained using stochastic variational inference (see Section~\ref{sec:svi}), the reparameterization trick~\citep{Kingma2014} is then used to recursively draw samples from the variational posterior:

\begin{align}
{\hat{f}}^{i,t}_{l} = &\;\widetilde{\mathbf{m}}_l \left(\left\lbrace{\hat{f}}^{i,t}_{l-1}, \textbf{x}^{i,t} \right\rbrace \right) + \nonumber\\
&\;\mathbf{\varepsilon}^{i,t}_l \odot \sqrt{\widetilde{\mathbf{S}}_l \left(\left\lbrace{\hat{f}}^{i,t}_{l-1},\textbf{x}^{i,t} \right\rbrace, \left\lbrace{\hat{f}}^{i,t}_{l-1},\textbf{x}^{i,t}\right\rbrace \right) }\;\;,
\end{align}

where $\mathbf{\varepsilon}^{i,t}_l\sim \mathcal{N}\left( \textbf{0}, \textbf{I}_{D_{\text{out}}} \right)$.

%Given univariate $p_1(x) = \norm(\mu_1, \sigma^2_1)$ and $p_2(x) = \norm(\mu_2, \sigma^2_2)$, the $D_\text{KL}$ term is computed as follows:
%
%$$
%D_\text{KL} \left(p_1(x) \| p_2(x)\right) = \frac{1}{2} 
%\left[
%\log\left(\frac{\sigma^2_2}{\sigma^2_1}\right) 
%-1 
%+ \frac{\sigma^2_1}{\sigma^2_2}
%+ \frac{(\mu_1 - \mu_2)^2}{\sigma^2_2}
%\right].
%$$

\section{Additional Detail on Experiments}

This appendix contains further information on the experimental evaluation provided in Section 5 of the main paper which was excluded due to space constraints.

\subsection{Mapping Between Fidelities for Synthetic Examples}

\begin{table}[h]
	\setlength{\tabcolsep}{14pt}
	\caption{Detail of synthetically-constructed functions used in experimental evaluation.}
	\begin{center}
		\begin{tabular}{lcc}
			\hlineB{3}
			\textsc{example}  &\textsc{fidelity}  &\textsc{function} \vspace{.1cm}\\
			\hline\\
			\textsc{linear-a} &\textsc{low} & $y_l\left(x\right) = \frac{1}{2}y_h\left(x\right) + 10\left(x - \frac{1}{2}\right) + 5$ \vspace{4pt}\\
			& \textsc{high} & $y_h\left(x\right) = \left(6x - 2\right)^2\sin\left(12x - 4\right)$ \\\\
			\textsc{linear-b} &\textsc{low} & $y_l\left(x\right) = 2y_h\left(x\right) + \left(x^3 - \frac{1}{2}\right)\sin\left(3x - \frac{1}{2}\right) + 4\cos\left(2x\right)$ \vspace{4pt}\\
			& \textsc{high} & $y_h(x) = 5x^2\sin(12x)$ \\
			\\
			\textsc{nonlinear-a} &\textsc{low} & $y_l\left(x\right) = \sin\left(8\pi x\right)$ \vspace{4pt}\\
			& \textsc{high} & $y_h\left(x\right) = \left(x - \sqrt{2}\right)\left(y_l\left(x\right)\right)^2$ \\\\
			\textsc{nonlinear-b} &\textsc{low} & $y_l\left(x\right) = \cos\left(15x\right)$ \vspace{4pt}\\
			& \textsc{high} & $y_h\left(x\right) = xe^{y_l\left(2x - .2\right)} - 1$\vspace{4pt}\\
			\hline
		\end{tabular}
	\end{center}
	\label{table:benchmark_detail}
\end{table}

\begin{figure}[h]
	\centering
	\begin{subfigure}[h]{0.5\textwidth}
		\centering
		\begin{tabular}{cc}
			\textsc{linear-a} & \textsc{linear-b}\\
			\includegraphics[width=110pt]{figures/functions/linearA.pdf} &
			\includegraphics[width=110pt]{figures/functions/linearB.pdf} \\
			\includegraphics[width=110pt]{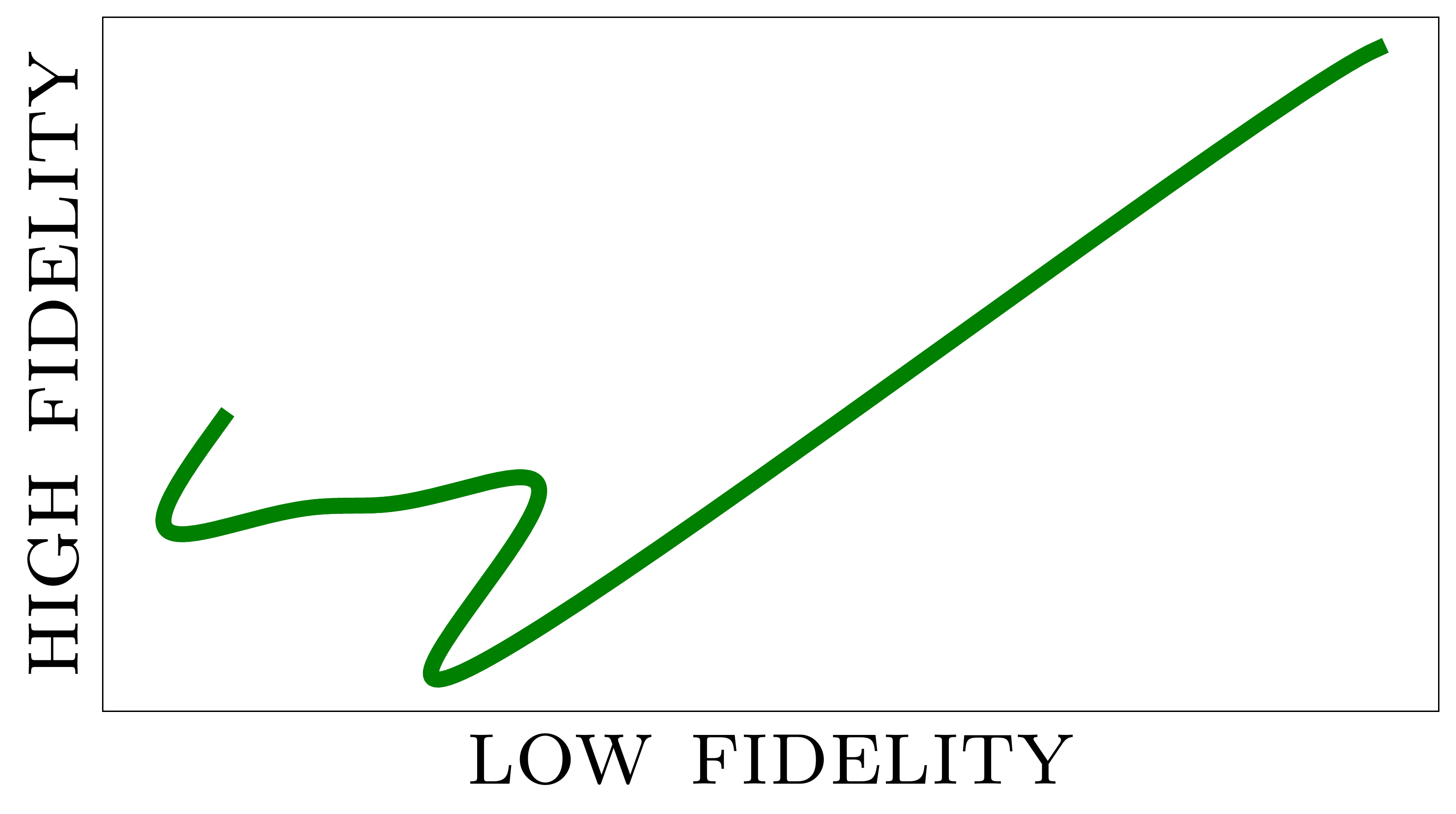} &
			\includegraphics[width=110pt]{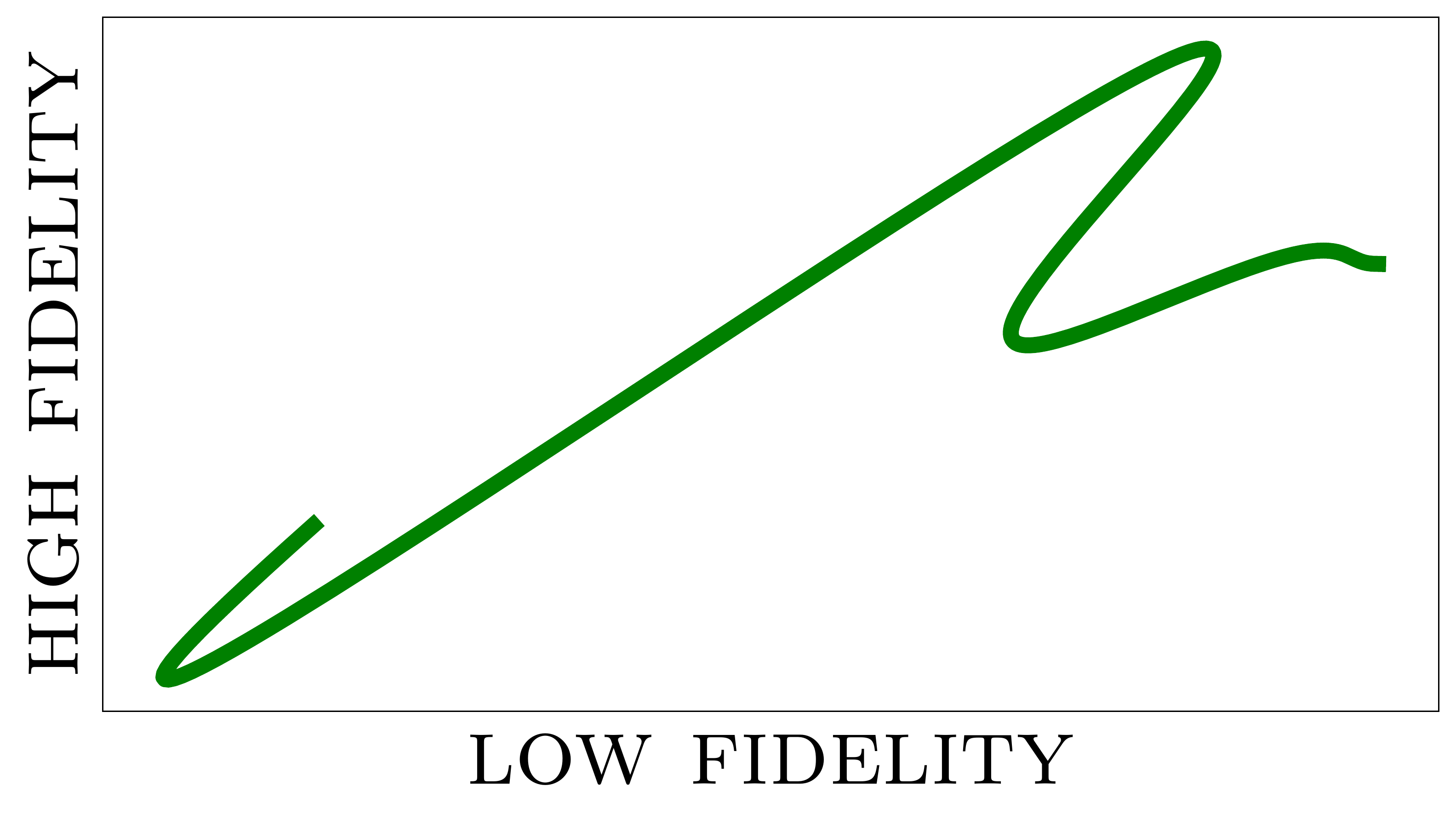}
		\end{tabular}
		\caption{Linear mapping.}
		\label{fig:mapping_a}
	\end{subfigure}%
	~ ~
	\begin{subfigure}[h]{0.5\textwidth}
		\centering
		\begin{tabular}{cc}
			\textsc{nonlinear-a} & \textsc{nonlinear-b}\\
			\includegraphics[width=110pt]{figures/functions/nonlinearA.pdf} &
			\includegraphics[width=110pt]{figures/functions/nonlinearB.pdf}\\
			\includegraphics[width=110pt]{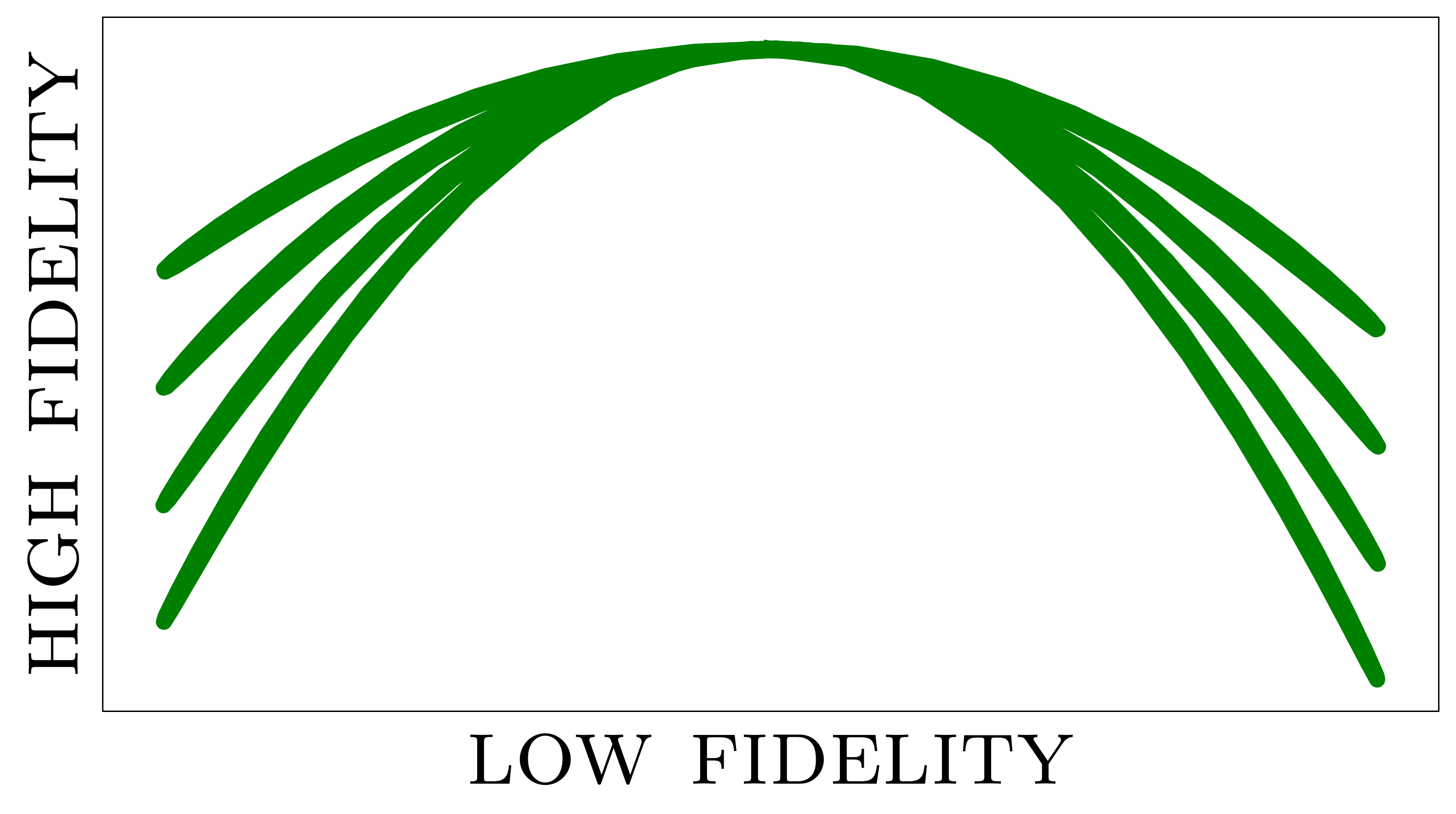} &
			\includegraphics[width=110pt]{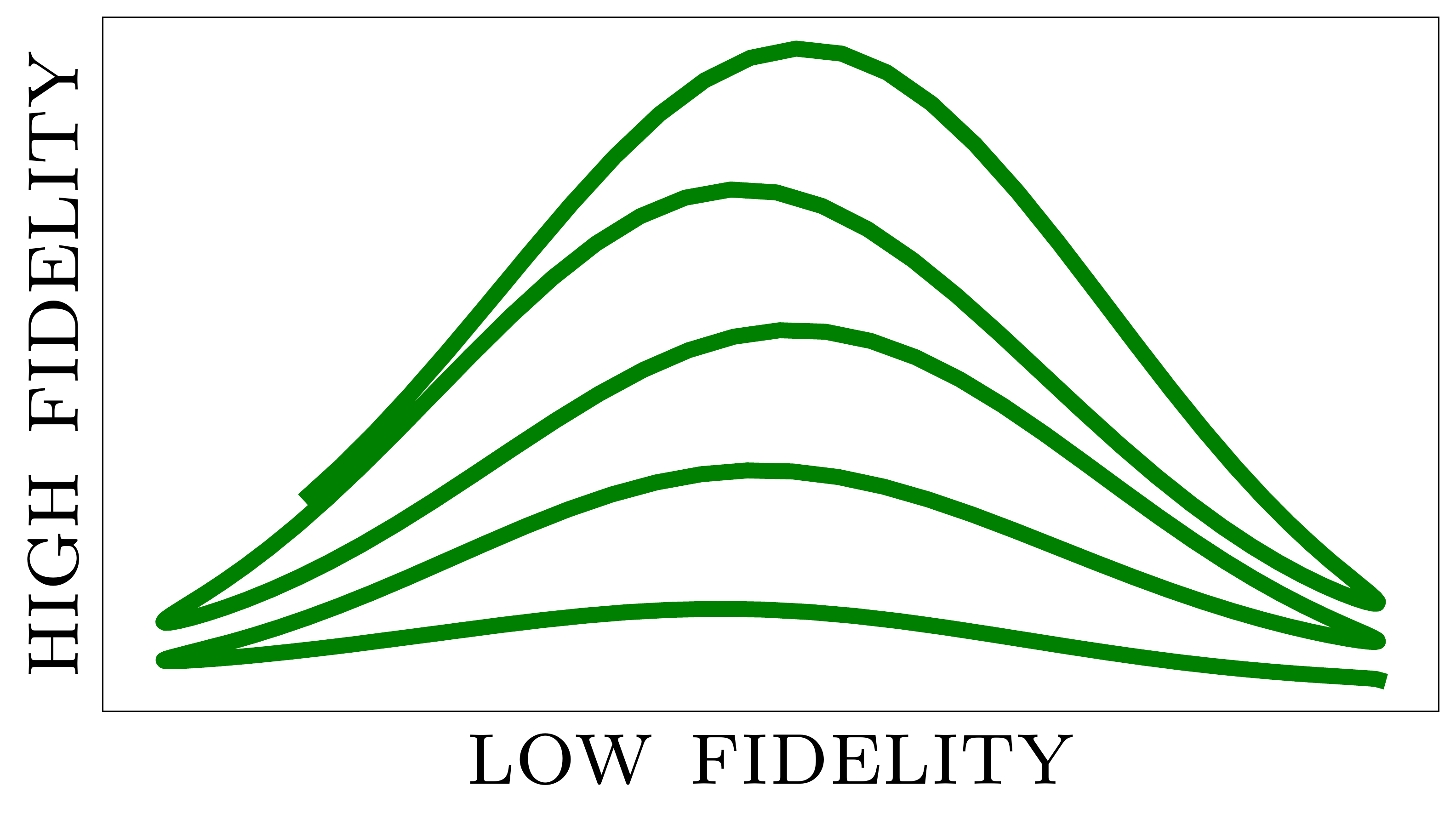}
		\end{tabular}
		\caption{Nonlinear mapping.}
		\label{fig:mapping_b}
	\end{subfigure}

	\caption{\textit{Top: }Synthetic multi-fidelity functions used for model comparison.
	\textit{Bottom: }Mapping between low and high-fidelity observations for same functions.}
	\label{fig:benchmarks_mapping}
\end{figure}

In the first experiment presented in Section \ref{sec:experiments}, we evaluated the performance of our model on four example functions, two having a linear mapping between fidelities and another two with nonlinear mappings; their precise definition is given in Table~\ref{table:benchmark_detail}.
The relationships between fidelities for these example functions are illustrated in Figure~\ref{fig:benchmarks_mapping}, where the bottom row shows the mapping from low-fidelity observations to their high-fidelity counterparts.
It is difficult to infer much useful information about the problem from simply observing these plots; however, the additional complexity of the two nonlinear examples is indicative of where the standard \ar model can be expected to perform badly.

In our evaluation, we observed that all methods worked best when the output values for all fidelities were scaled down, particularly for ensuring convergence in the optimization procedure.
To this end, in the experiments we scale down the original functions by a constant scaling factor while still preserving the relationship between fidelities in their original formulation.
%The values for mean squared error given in the main paper are then computed by first scaling up predictions back to their original scale.

\subsection{Specification of Benchmark Problems}

In Section~\ref{sec:exp_benchmarks} of the main paper, we evaluated the performance of our model on a set of five benchmark problems that are widely used in the literature for evaluating the effectiveness of multi-fidelity methods.
The specification of each problem is given below:

\begin{itemize}

	\item \textsc{currin}
	
	The \textsc{currin} function is a two-dimensional problem that is commonly featured in works related to simulating computer experiments, with input domain $\xvect\in\left[0,1\right]^2$.
	The high-fidelity variation of this function is given by:
	
	\begin{align*}
		y_h\left(\xvect\right) = \left[1 - \exp\left(-\frac{1}{2x_2}\right)\right]\frac{2300x_1^3 + 1900x_1^2 + 2092x_1 + 60}{100x_1^3 + 500x_1^2 + 4x_1 + 20} \text{,}
	\end{align*}
	
	\noindent whereas the low-fidelity alternative is given by:
	
	\begin{align*}
		y_l\left(\xvect\right) = & \frac{1}{4}\left[y_h\left(x_1 + 0.05, x_2 + 0.05\right) + y_h\left(x_1 + 0.05, \max\left(0, x_2 - 0.05\right)\right)\right] +\\
		&\frac{1}{4}\left[y_h\left(x_1 - 0.05, x_2 + 0.05\right) + y_h\left(x_1 - 0.05, \max\left(0, x_2 - 0.05\right)\right)\right] \text{;}
	\end{align*}
	
	\item \textsc{park}
	
	The \textsc{park} function is a four-dimensional problem where all inputs lie in the range $\left[0,1\right]$.
	High-fidelity observations are evaluated as:
	
	\begin{align*}
		y_h\left(\xvect\right) = \frac{x_1}{2}\left[\sqrt{1 + \left(x_2 + x_3^2\right)\frac{x_4}{x_1^2}} - 1\right] + \left(x_1 + 3x_4\right)\exp\left[1 + \sin\left(x_3\right)\right] \text{,}
	\end{align*}
	
	\noindent while low-fidelity observations are obtained using:
	
	\begin{align*}
		y_l\left(\xvect \right) = \left[1 + \frac{\sin\left(x_1\right)}{10}\right]y_h\left(\xvect\right) - 2x_1 + x_2^2 + x_3^2 + 0.5 \text{;}
	\end{align*}
	
	\item \textsc{borehole}
	
	The \textsc{borehole} example is a two-level physical model that simulates water flow through a borehole, and depends on eight input parameters.
	The input domain is constrained to lie in the following regions: $x_1\in \left[0.05,0.15\right]$, $x_2\in\left[100,50000\right]$, $x_3\in\left[63070,115600\right]$, $x_4\in\left[990,1110\right]$, $x_5\in\left[63.1,115\right]$, $x_6\in\left[700,820\right]$, $x_7\in\left[1120,1680\right]$, $x_8\in\left[9855,12045\right]$.
	The high-fidelity simulation for this model is given by:
	
	\begin{align*}
		y_h\left(\xvect\right) = \frac{2\pi x_3\left(x_4-x_6\right)}{\log\left(x_2/x_1\right)\left(1 + \frac{2x_7x_3}{\log\left(x_2/x_1\right)x_1^2x_8}\right) + \frac{x_3}{x_5}} \text{,}
	\end{align*}
	
	\noindent while the low-fidelity variant is evaluated as:
	
	\begin{align*}
		y_l\left(\xvect\right) = \frac{5x_3\left(x_4 - x_6\right)}{\log\left(x_2/x_1\right)\left(1.5 + \frac{2x_7x_3}{\log\left(x_2/x_1\right)x_1^2x_8}\right) + \frac{x_3}{x_5}} \text{;}
	\end{align*}
	
	\item \textsc{branin}
	
	The three-level \textsc{branin} function is taken from the specification given in \cite{Perdikaris2017}, where the two-dimensional input lies in the range $\left[-5,10\right]\times\left[0,15\right]$.
	The three tiers are given by:
	
	\begin{align*}
		y_h\left(\xvect\right) = \left(\frac{-1.275x_1^2}{\pi^2} + \frac{5x_1}{\pi} + x_2 - 6\right)^2 + \left(10 - \frac{5}{4\pi}\right)\cos\left(x_1\right) + 10 \text{,}
	\end{align*}
	\begin{align*}
		y_m\left(\xvect\right) = 10\sqrt{y_h\left(\xvect - 2\right)} + 2\left(x_1 - 0.5\right) - 3\left(3x_2 - 1\right) -1 \text{, and}
	\end{align*}
	\begin{align*}
		y_l\left(\xvect\right) = y_m\left(1.2\left(\xvect + 2\right)\right) - 3x_2 + 1 \text{;}
	\end{align*}
	
	\item \textsc{hartmann-{\small 3}d}
	
	Finally, the three-level \textsc{hartmann-{\small 3}d} example follows the specification provided in \cite{Kandasamy2016}, whereby the three-dimensional input lies in the domain $\left[0,1\right]^3$.
	The evaluation of observations with fidelity $t$ is given by:
	
	\begin{align*}
		y_t\left(\xvect\right) = \sum_{i=1}^4\alpha_i\exp\left(-\sum_{j=1}^3A_{ij}\left(x_j - P_{ij}\right)^2\right) \text{,}
	\end{align*}
	
	\noindent where 
	
		$$
			A =
			\begin{bmatrix}
			    3	   & 10 & 30 \\
			    0.1   & 10 & 35 \\
			    3      & 10 & 30 \\
			    0.1	  & 10 & 35
			\end{bmatrix} \quad\quad \text{and}\quad\quad 
			P = \begin{bmatrix}
				    0.3689	& 0.1170 & 0.2673 \\
				    0.4699  & 0.4387 & 0.7470 \\
				    0.1091  & 0.8732 & 0.5547 \\
				    0.0381  & 0.5743 & 0.8828
				\end{bmatrix}  \text{.}
		$$ 
		
		\noindent The vector $\pmb{\alpha}$ is initially set to $\left[1.0,1.2,3.0,3.2\right]^\top$ and is updated to $\pmb{\alpha}_t = \pmb{\alpha} + (3 - t)\pmb{\delta}$ for lower fidelities, where $\pmb{\delta} = \left[0.01,-0.01,-0.1,0.1\right]^\top$.
\end{itemize}

\subsection{Configuration of Competing Models}

In this final section, we elaborate on the configuration and optimization strategies used for the competing techniques in Section~\ref{sec:experiments} of the paper.
%Complementing the details given in the main paper regarding the configuration of \mfdgp, the upcoming discussion is intended to guarantee the reproducibility of experiments featured in this work.
%The models used in this work have been re-implemented and integrated in an open-source library.\footnote{Obfuscated to preserve anonymity.}

\begin{itemize}
	\item \ar \citep{Kennedy2000}
	
	The \ar model is implemented as per the original specification presented by \cite{Kennedy2000}.
	We opt for this formulation instead of the procedure detailed in \cite{LeGratiet2014} since the latter is more cumbersome to adapt to non-nested input structures, whereas this constraint does not apply to the former.
	We assign independent noise parameters to each fidelity, which are jointly optimized with the kernel hyperparameters and scaling factors in a single call to the optimization procedure.

	\item \nargp \citep{Perdikaris2017}
	
	For the \nargp model, we adopt the same optimization strategy considered by \cite{Perdikaris2017} in their evaluation.
	In particular, individual \gps are used for modeling the data at each fidelity level, and these are optimized sequentially in isolation.
	We optimize the kernel parameters for the \gps at each layer using a two-step procedure which was applied in the original implementation provided by the authors - the optimization is first carried out with fixed noise variance, after which this parameter is also freed and all parameters are adapted jointly.
	
	\item \deepmf \citep{Raissi2016}
	
	One of the challenges associated with the \deepmf model is in selecting an appropriate deterministic nonlinear transformation to be applied to the input data.
	Given that there is no straightforward approach for deciding how to configure this component of the model, in our evaluation we use the two-layer neural network with sigmoid activation functions reported in the original presentation of the model given by \cite{Raissi2016}.
	The process noise is shared between fidelities.
	No pre-existing code was found for this model.
\end{itemize}

\end{document}